\documentclass[letterpaper]{article} % DO NOT CHANGE THIS
\usepackage{aaai24}  % DO NOT CHANGE THIS
\usepackage{times}  % DO NOT CHANGE THIS
\usepackage{helvet}  % DO NOT CHANGE THIS
\usepackage{courier}  % DO NOT CHANGE THIS
\usepackage[hyphens]{url}  % DO NOT CHANGE THIS
\usepackage{graphicx} % DO NOT CHANGE THIS
\usepackage{natbib}  % DO NOT CHANGE THIS AND DO NOT ADD ANY OPTIONS TO IT
\usepackage{caption} % DO NOT CHANGE THIS AND DO NOT ADD ANY OPTIONS TO IT
\usepackage{algorithm}
\usepackage{algorithmic}

\usepackage{epsfig}
\usepackage{amsmath}
\usepackage{amssymb}
\usepackage{xcolor}
\usepackage{lipsum}
\usepackage{color}
\definecolor{turquoise}{cmyk}{0.65,0,0.1,0.3}
\definecolor{purple}{rgb}{0.65,0,0.65}
\definecolor{dark_green}{rgb}{0, 0.5, 0}
\definecolor{orange}{rgb}{0.8, 0.6, 0.2}
\definecolor{red}{rgb}{0.8, 0.2, 0.2}
\definecolor{darkred}{rgb}{0.6, 0.1, 0.05}
\definecolor{blueish}{rgb}{0.0, 0.3, .6}
\definecolor{light_gray}{rgb}{0.7, 0.7, .7}
\definecolor{pink}{rgb}{1, 0, 1}
\definecolor{greyblue}{rgb}{0.25, 0.25, 1}
\usepackage{booktabs}
\usepackage{color, colortbl}
\usepackage{multirow}
\usepackage{color}
\definecolor{turquoise}{cmyk}{0.65,0,0.1,0.3}
\definecolor{purple}{rgb}{0.65,0,0.65}
\definecolor{dark_green}{rgb}{0, 0.5, 0}
\definecolor{orange}{rgb}{0.8, 0.6, 0.2}
\definecolor{red}{rgb}{0.8, 0.2, 0.2}
\definecolor{darkred}{rgb}{0.6, 0.1, 0.05}
\definecolor{blueish}{rgb}{0.0, 0.3, .6}
\definecolor{light_gray}{rgb}{0.7, 0.7, .7}
\definecolor{pink}{rgb}{1, 0, 1}
\definecolor{greyblue}{rgb}{0.25, 0.25, 1}
\usepackage{newfloat}
\usepackage{listings}
\DeclareCaptionStyle{ruled}{labelfont=normalfont,labelsep=colon,strut=off} % DO NOT CHANGE THIS
\floatstyle{ruled}
\newfloat{listing}{tb}{lst}{}
\floatname{listing}{Listing}
\title{ICAR: Image-based Complementary Auto Reasoning}
\author {
    Xijun Wang\textsuperscript{\rm 1},
    Anqi Liang\textsuperscript{\rm 2},
    Junbang Liang\textsuperscript{\rm 2},
    Ming Lin\textsuperscript{\rm 1,2},
    Yu Lou\textsuperscript{\rm 2},
    Shan Yang\textsuperscript{\rm 2}
}
\affiliations {
    \textsuperscript{\rm 1} University of Maryland, College Park, 
    \textsuperscript{\rm 2} Amazon\\
    \{xijun, lin\}@umd.edu, \{lianganq, junbangl, ylou, ssyang\}@amazon.com
}
\usepackage{bibentry}
\begin{document}

%\maketitle

\twocolumn[{%
\renewcommand\twocolumn[1][]{#1}%
\maketitle
\vspace{-5em}
\begin{center}
    \centering
    \captionsetup{type=figure}
   \includegraphics[width=\textwidth,height=5cm]{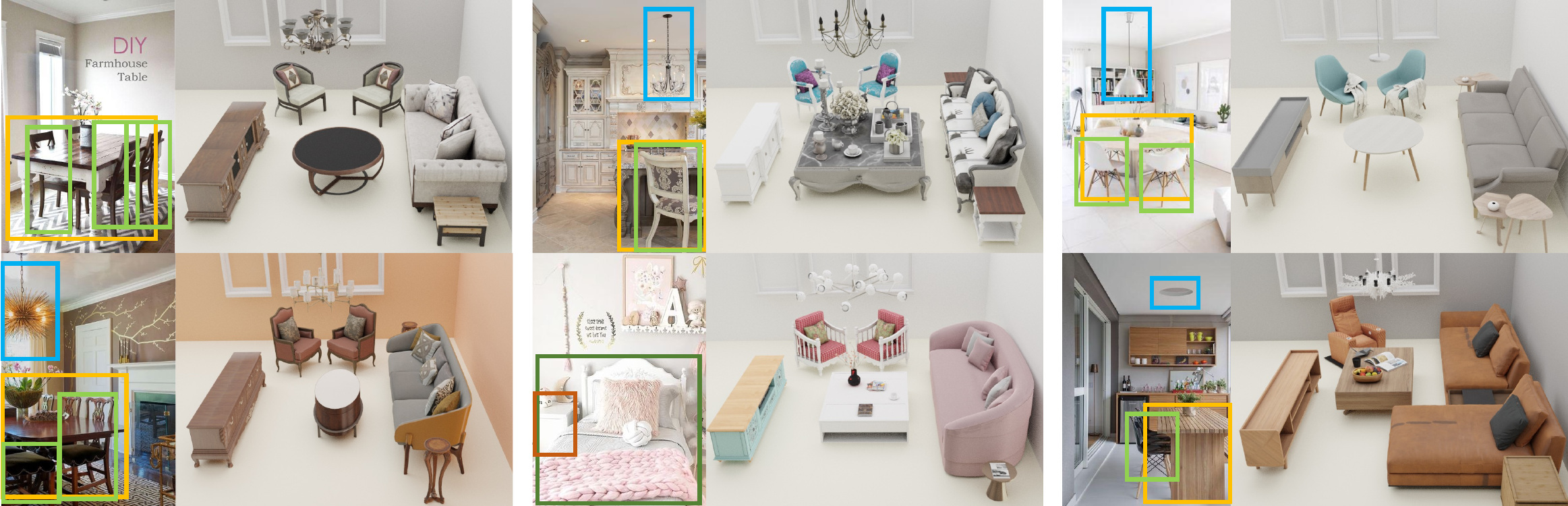}
    \vspace{-1.5em}
    \captionof{figure}{\textbf{Cover Image:} We present a new self-supervised model for scene-aware, visually compatible 
    %cross-domain 
    object retrieval tasks. In this example, given an inspirational home scene image (sampled from STL-home\cite{kang2019complete} column 1, 3, 5) with a pool of objects (3D-FRONT\cite{fu20213d}) from an unseen domain, our model auto-regressively retrieves a set of stylistically compatible items (column 2, 4, 6).
    %We are also one of the first to demonstrate the capability of generalizing to unseen domain for the scene-aware cross-domain CIR task.
    % We construct our compatibility learning model based on two major influencing factors on this task : {\em cross-domain similarity} and {\em complementary set compatibility}.
    % We propose a category-aware {\em Flexible Bidirectional Transformer} model to automatically generate a set of visually compatible items.  This method is able to achieve an impressive improvement up to {\bf 9.6\%} in Fill-In-The-Blank (FITB) score~\cite{han2017learning} and {\bf 31.8\%} in Style FID (SFID) for furniture, as shown above. 
    }
\label{fig:teaser}
\end{center}%
% \vspace{-1em}
}]

\begin{abstract}
Scene-aware Complementary Item Retrieval (CIR) is a challenging task which requires to generate a set of compatible items across domains. 
Due to the subjectivity, it is difficult to set up a rigorous standard for both data collection and learning objectives. To address this challenging task, we propose a {\em visual compatibility} concept, composed of {\em similarity} (resembling in color, geometry, texture, and etc.) and {\em complementarity} (different items like table vs chair completing a group). Based on this notion, we propose a {\em compatibility learning} framework, a category-aware Flexible Bidirectional Transformer (FBT), for visual ``scene-based set compatibility reasoning'' with the {\em cross-domain} visual similarity input and auto-regressive complementary item generation. We introduce a ``Flexible Bidirectional Transformer (FBT),'' consisting of an encoder with flexible masking, a category prediction arm, and an auto-regressive visual embedding prediction arm. And the inputs for FBT are cross-domain visual similarity invariant embeddings, making this framework quite generalizable. Furthermore, our proposed FBT model learns the inter-object compatibility from a large set of scene images in a self-supervised way. 
%Furthermore, to measure set visual compatibility quantitatively, we propose a new CIR metric,  Style-FID (SFID). We validate via user study that our new SFID metric correlates well with human perception. 
Compared with the SOTA methods, this approach achieves up to {\bf 5.3\%} and {\bf 9.6\%} in FITB score and {\bf 22.3\%} and {\bf 31.8\%} SFID improvement on fashion and furniture, respectively. 
\end{abstract}

%Scene-aware Complementary Item Retrieval (CIR) is a challenging task which requires to generate a set of compatible items. Due to the subjectivity, it is difficult to set up a rigorous standard for both data collection and learning objectives. To solve this challenging task, we propose the visual compatibility concept which composes of similarity (color, geometry, texture, and etc.) and complementarity (different items like table vs chair). Based on this concept. we put forward a compatibility learning framework, a category-aware Flexible Bidirectional Transformer (FBT), for visual scene-based set compatibility reasoning with the cross-domain visual similarity input and auto-regressive complementary item generation. For the Flexible Bidirectional Transformer (FBT), in which we introduce a bidirectional Transformer encoder with flexible masking, a category prediction arm, and an auto-regressive visual embedding prediction arm. And the input for FBT are cross-domain visual similarity invariant embeddings, whcih makes our frame work has a excellent generalization. Furthermore, our proposed FBT model learns the inter-object compatibility from a large set of scene images in a self-supervised way. Compared with the SOTA method, our method achieves up to 5.3% FITB score and 22.3% SFID improvement on fashion, and up to 9.6% FITB score improvement and 31.8% SFID improvement on furniture. 

\section{Introduction}
% Today technologies have reinvigorated how people shop online. 
Online shopping catalogs provide great convenience, such as searching and comparing similar items. 
% While digital catalog provides a fast way of e-shopping, 
However, when customers can compare similar items, they often miss the browsing of complementary items in the e-shopping experience. Millions of online images offer a new opportunity to shop with inspirational home decoration ideas or outfit matching.  But, to retrieve stylistically compatible products from these online images for set matching can be an overwhelming process.% even for human. 
The ability to recommend visually complementary items becomes especially important, when shopping for home furniture and clothings.
% In the meanwhile, it is very challenging to abstract or describe the essence of a good home design or a matching fashion outfit. 
The subjectivity makes the visual compatibility even more difficult to model computationally.

\begin{figure}
\centering
\includegraphics[width=1.0\columnwidth]{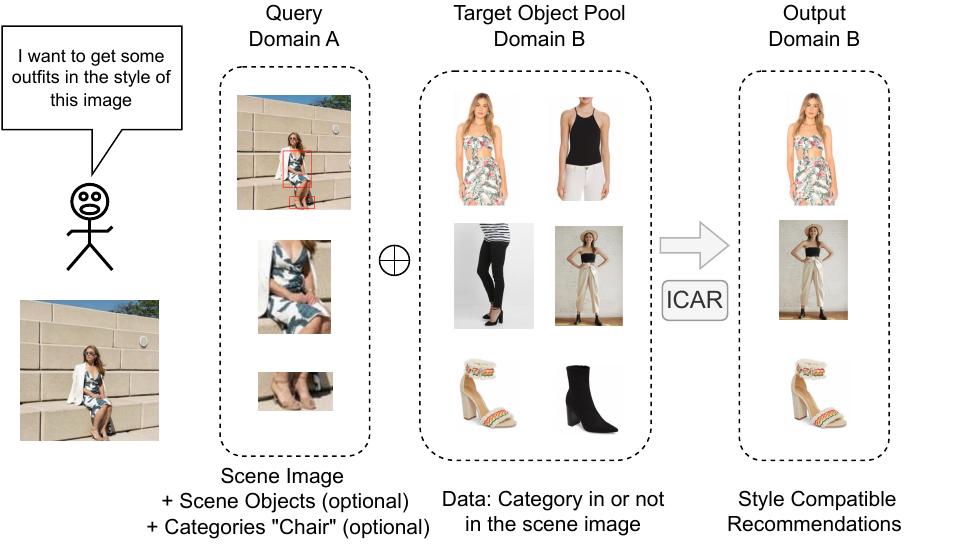}
\vspace{-1.5em}
\caption{\textbf{Scene-aware Complementary Item Retrieval Task Illustration.} Given a query scene image, (optional) scene objects and item categories, the task goal is to generate a cross-domain set of stylistically compatible items. 
}
\label{fig:intro}
\vspace{-1.5em}
\end{figure}

In this work, we aim to address the {\em visual scene-aware Complementary Item Retrieval} (CIR)~\cite{sarkar2022outfittransformer,kang2019complete} task.
In this task (as shown in Figure~\ref{fig:intro}), we attempt to model human's ability to select a set of objects from cross-domain pools, given a scene image, objects in the scene, and object categories. Therefore, we propose a {\em visual compatibility} concept, consisting of two key elements: 
{\em similarity} and {\em complementarity}.
%Scene-aware CIR is a challenging task as it requires both visual similarity and complementarity modeling.
Visual similarity and complementarity, however, can contradict each other sometimes.
Items that look similar (color, geometry, texture, and etc.) may not be complementary (different items like dinner table vs sofa) when putting them into a set. Items that complement each other do not necessarily look similar (e.g. an outfit set in contrasting colors). 
%To enable cross-domain generation, visual similarity is required to be modeled.
%It composes two parts: similarity learning and complementary learning. To some degree these two aspects contradicts each other. 
The ambiguous definition for visual complementarity is a major challenge. 
This ambiguity makes it difficult to rigorously define an objective and creates extra challenge for collecting such datasets, when designing a data-driven method. 
% What makes it even more challenging % and worth solving 
% is the problem of conditional scene-based Complementary Item Retrieval. 
% In Conditional Complementary Item Retrieval, one needs to take into account of personal preferences, multi-modal input, etc.. 

To address these issues, we first propose a ``compatibility leaning'' framework to model the visual {\em similarity} and {\em complementarity}. To the best of our knowledge, we are among the first to show qualitatively that our model based on this framework can generalize to unseen domains (where the model is not trained with, as shown in Figure~\ref{fig:teaser} and Figure~\ref{fig:visual}).
%End-to-end learning has the advantages of avoiding error accumulation but it suffers from poor convergence for complex objectives. 
For the scene-based CIR task, it's complex to learn both the cross-domain similarity and complementarity. Therefore, we use cross-domain visual similarity invariant embeddings in our framework.
%, \XW{which will make the training process tedious since it needs to train from the scratch every time.}
Many previous CIR works~\cite{han2017learning,kang2019complete} also start from some types of learned embedding.
But failing to model the visual similarity creates extra complexity for the complementary learning.
Secondly, we propose to use self-supervised learning for visual complementarity reasoning by introducing an auto-regressive transformer based architecture.
Given the difficulty to define style complementarity mathematically, we propose a solution based on the assumption that the items exist in the inspirational scene images are compatible with each other. 
%Therefore, we extract the compatibility information self-supervisely with no extra label needed.

Built upon the aforementioned premises, we present a novel self-supervised transformer-based learning framework (overview shown in Fig.~\ref{fig:model_overview}).
Our model effectively learns both the similarity and complementarity between a set of items. 
Our model does not require extra complementary labels. 
In addition, compared to the prior work that models complementary items as {\em pairs} or a {\em sequence of items}, we model them as {\em unordered sets}. 
%Our learning framework is general and extensible to solve both the conditional and scene-based Complementary Item Retrieval (CIR) task. 
We carefully design our compatibility leaning model. 
First, we ensure that the learned embedding both contains and extracts all the necessary information for the compatibility learning. 
Second, we make full use of Transformer's ability in reasoning about the interactions between these learned embeddings. 
%But vanilla Transformer is not suitable for 
To model flexible-length unordered set generation with cross-domain retrieval, we propose a new Flexible Bidirectional Transformer (FBT). 
In this FBT model, we model the unordered set generation using random shuffle and masking technique.
In addition, we introduce a category prediction arm and a cross-domain retrieval arm to the transformer encoder. 
The added category prediction branch helps the model to reason about the complementary item types. 
%We demonstrate in our ablation study that our method learns well about the item category distribution. 
As pointed out in \cite{vasileva2018learning}, the category embedding representation of each item  carries the notion of both similarity and complementarity. 
Our proposed model, compared with prior work that also models the multi-item compatibility using neural networks, such as works by~\cite{li2017mining}, ~\cite{han2017learning} and ~\cite{sarkar2022outfittransformer}, does not require the partial sets to be given. 

We validate our method on the CIR benchmark datasets include, Shop The Look (STL)~\cite{kang2019complete}, Exact Street2Shop~\cite{hadi2015buy}  and DeepRoom~\cite{gadde2021detail}. 
Our method consistently outperforms the state-of-the-art methods. 
%Together, we present an extensive ablation study.
%for each new parts in our two-stage learning framework.
More importantly, we notice that most of the CIR prior work are evaluated via the Fill-In-The-Blank (FITB)~\cite{han2017learning} metric or human in the loop.
The FITB metric can reflect the model's ability in cross-domain retrieval but it does not measure the complementarity as a set.
Human in the loop evaluation, however, is both limited to scale and biases, if not conducted thoughtfully.
To address these issues, we propose a new CIR evaluation metric: ``Style Frechet Inception Distance'' (SFID) (see supplementary for details).
%as none of the previous metrics evaluate the set compatibility. 
%\JL{extend this sentence to a paragraph. why do we need this? how it can help us evaluate?}
%We also demonstrate that our model can be easily extended to personalized complementary item recommendation.

In summary, the key contributions of this work include: 
\begin{itemize}
  \item {\bf Visual compatibility} is defined based on {\em similarity} and {\em complementarity} for the Scene-aware Complementary Item Retrieval (CIR) task and a new {\em compatibility learning framework} is designed to solve this task.
  \item For the {\bf compatibility learning framework}, a category-aware {\em Flexible Bidirectional Transformer (FBT)} is introduced for visual scene-based set compatibility reasoning with the {\em cross-domain} visual similarity input and auto-regressive complementary item generation.
  % that focuses on the two major influencing factors in this task: cross-domain visual similarity and reason the complementary set visual compatibility;
  % \item a {\em cross-domain retrieval } embedding module for visual similarity learning;
  %\item a category-aware {\em Flexible Bidirectional Transformer (FBT)} for visual scene-based set compatibility reasoning, and the corresponding set visual complementary generation metric, {\em Style Frechet Inception Distance (SFID)}.
\end{itemize}
% 1) a new learning framework that focuses both on {\em similarity and compatibility};
% 2) a {\em cross-domain retrieval } embedding module for similarity learning;
% 3) a category-aware {\em Flexible Bidirectional Transformer (FBT)} for visual scene-based set compatibility reasoning, and the corresponding set visual complementary generation metric, {\em Style Frechet Inception Distance (SFID)}.

\section{Related Work}
% The visual cross-domain scene-aware Complementary Item Retrieval task is related to both similarity and compatibility learning. 

\paragraph{Visual Similarity Learning}
Visual similarity learning has been a main computer vision topic. The goal is to mimic human's ability in finding visually similar objects or scenes. %From the definition of similarity, this task can be further divided into, image object classification~\cite{chen2017rethinking,simonyan2014very,krizhevsky2017imagenet,he2016deep,xie2017aggregated} - finding objects in the same category, object detection~\cite{zou2019object,felzenszwalb2010object,girshick2014rich,girshick2015fast,he2017mask,redmon2017yolo9000} - identifying similar objects across different scenes, 
This is particularly studied in image retrieval~\cite{el2021training,radenovic2018fine,teh2020proxynca++,cheng2021fashion} - finding images with a certain definition of similarity and so on.
%Our cross-domain CIR task is closely related to image retrieval. 
Prior to retrieving similar clothing, researchers also studied how to detect and segment clothing from real-life images~\cite{yamaguchi2012parsing,yang2011real,gallagher2008clothing}. 
With clothing detection or segmentation, similar clothing retrieval is explored via style analysis~\cite{hsiao2017learning,kiapour2014hipster,simo2016fashion,yu2012dressup,yamaguchi2013paper,kiapour2014hipster,di2013style}. 
In the meantime, Liu ~\cite{liu2012street,liu2012hi} pioneer ways to do cross-domain retrieval which retrieves  similar clothing from real-life images and targets images from a different domain, such as well-staged product images. 
Recent fashion retrieval tasks can be further categorized based on the input information, such as images~\cite{kalantidis2013getting,liu2016deepfashion,simo2016fashion,zhai2017visual,hadi2015buy,tran2019searching}, clothing attributes~\cite{ak2018learning,di2013style}, videos~\cite{cheng2017video2shop}.

\noindent\textbf{Visual complementarity Learning} 
Visual complementarity learning, unlike visual similarity learning, is much more ambiguous. 
%In the fashion and home furniture recommendation applications, the ability to recommend complementary items is critical. 
There are a couple of research directions: pairwise complementary item retrieval~\cite{veit2015learning,song2017neurostylist,vasileva2018learning,mcauley2015image,lin2020fashion,taraviya2021personalized}, set complementary prediction (no cross domain retrieval)~\cite{tangseng2017recommending,li2017mining,han2017learning, hsiao2018creating,tangseng2017recommending,shih2018compatibility,li2020hierarchical}, set complementary item retrieval~\cite{hu2015collaborative,huang2015cross,liu2012hi}, personalized set complementary item prediction (requires user input)~\cite{taraviya2021personalized,chen2019pog,li2020hierarchical,su2021complementary,zheng2021collocation,guan2022bi,guan2022personalized} and multi-modal complementary item prediction~\cite{guan2021multimodal}. 
All these prior work focus on feature representation learning. 
%while still use CNN for pixel to feature projection. 
Another line of works~\cite{chen2015deep,song2017learning,tan2019learning,lin2020fashion} focus on learning multiple sub-embedding based on different properties for both similarity and compatibility. More recently, Transformer~\cite{vaswani2017attention} has demonstrated strong performance across various natural language processing and computer vision tasks. 
Kang ~\cite{kang2019complete} proposes to use CNN visual classification features and attention mechanism.
Later on, Sarkar ~\cite{sarkar2022outfittransformer} uses Transformer and CNN-based image classification features for compatibility learning.
Similarly, Chen ~\cite{chen2019pog} applies Transformer together with CNN image classification feature to learn a mapping between user picked item pool and a set of most compatible items from that pool. 
%Kang ~\cite{kang2019complete}, Sarkar ~\cite{sarkar2022outfittransformer} and Chen ~\cite{chen2019pog} are closest to ours in terms of applying Transformer for scene-based visual complementarity learning.
Unlike all the work above, %we build our two-stage model to separate the visual similarity and compatibility learning. 
we build our visual compatibility model which focuses on both similarity and complementarity.
%Besides, we innovate and propose the Flexible Bidirectional Transformer (FBT) for un-ordered scene and category-aware item set compatibility learning.

\noindent\textbf{Learning Framework}
% Instead of solving the problem end-to-end, we propose the two-stage modeling. 
% We are not the first to use this technique. 
Many researchers have studied and explored building a cascaded learning framework. 
The cascaded method here means learning how to encode the data then modeling the statistics of this encoding. 
%This includes two-stage VAE~\cite{dai2019diagnosing} and many recent text to image generation work like VQGAN~\cite{esser2021taming}, DALL-E~\cite{ramesh2021zero} and DALL-E2~\cite{ramesh2022hierarchical}.
Many of the methods proposed for CIR task can also be categorized as two-stage models. 
But almost all of them use the image classification training target as the first-stage feature extractor~\cite{han2017learning,sarkar2022outfittransformer,kang2019complete,chen2019pog}.
Taraviya ~\cite{taraviya2021personalized} propose a two-stage model for personalized pairwise complementary item recommendation where they learn a feature embedding specially designed for customer preferences in their first stage. 
In our compatibility learning, we set cross-domain visual similarity embedding as input, and design FTB for complementary set generation. We show empirically that the visual similarity feature, compared to image classification learned features, are better suited for CIR. Our design makes our model surpass the prior work in scene aware CIR task. 

% %  move to supp
% \noindent\textbf{Visual complementarity Learning} 
% Veit ~\cite{veit2015learning} proposes Siamese CNN for pairwise style embedding learning. 
% Han ~\cite{han2017learning} extends it from pair to set and models the set as an ordered sequence using LSTM.

% Veit ~\cite{veit2017conditional} proposes to learn two more sub-embedding to help with compatibility learning.
% Further Vasileva ~\cite{vasileva2018learning} improves the learned embedding by emphasizing the dis-entanglement on item types. 
% Similar to Han ~\cite{han2017learning}, Cucurull ~\cite{cucurull2019context} applies graph neural network for compatibility learning and CNN with image classification as target for first stage feature extractor. 
\section{Method}
\begin{figure*}[t]
\vspace{-1mm}
\centering
\includegraphics[width=1.0\textwidth]{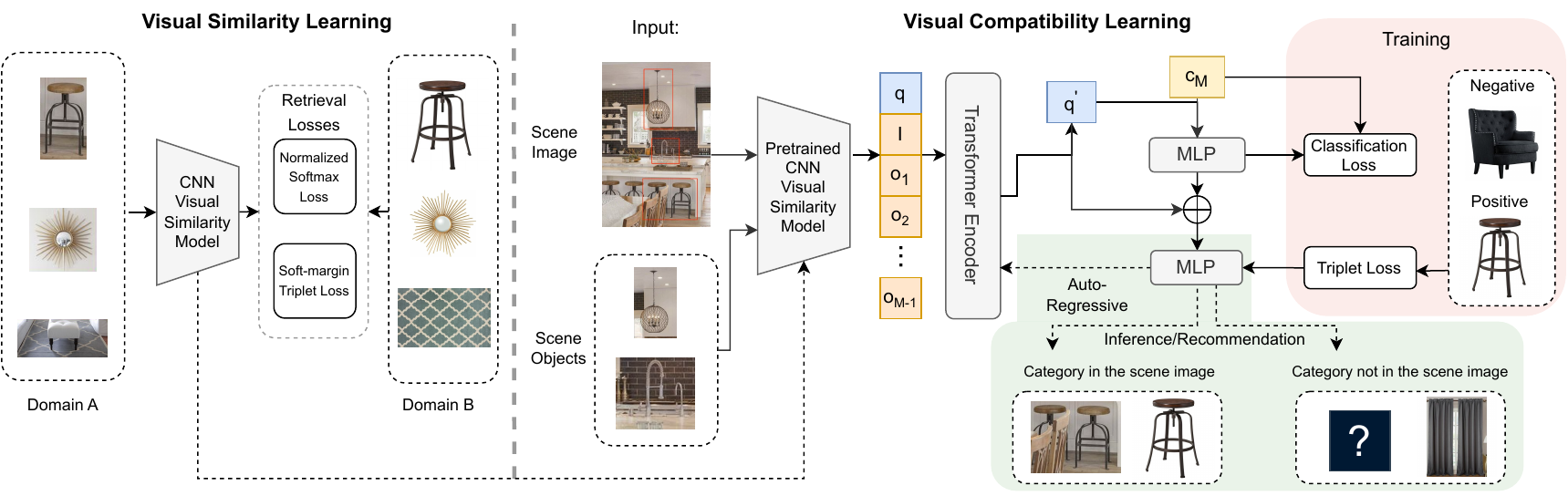}
\caption{\textbf{ICAR Model Overview.} In similarity learning, we apply a CNN-based model\cite{jun2019combination} to learn the visual similarity features across two domains. 
The learned features are required for both complementary reasoning in the complementarity learning and the cross-domain retrieval. 
With the learned features, in the complementarity learning, we propose a Flexible Bidirectional Transformer (FBT) model to learn the multi-object visual compatibility.
}
\label{fig:model_overview}
\vspace{-3mm}
\end{figure*}
%Next we introduce our two-stage learning framework for the scene-based Complementary Item Retrieval task.
\paragraph{Problem Statement}
Given a scene image $\mathcal{I}$, a set of unordered objects $\mathcal{O}=\{o_i\}_{i=0}^{N}, o_i\in\mathcal{D}_A$ in the scene and a set of unordered object categories $\mathcal{C}=\{c_i\}_{i=0}^L$, the problem is to retrieve cross-domain a set of complementary objects $\mathcal{X}=\{x_i\}_{i=0}^L, x_i\in\mathcal{D}_B$. 
This generated set of objects needs to be both visually compatible with each other and of visually similar style to the input scene image $\mathcal{I}$. 
Here we use $\mathcal{D}_A$ and $\mathcal{D}_B$ to denominate the two different visual domains, $L$ to represent the number of objects to retrieve during inference and $N$ is the number of scene objects. 
The difference between the two domains $\mathcal{D}_A$ and $\mathcal{D}_B$ can be quantified as the Fr\'echet distance $\mathcal{F}$ larger than a certain threshold $\theta$.
%Our goal in this paper is to retrieve cross domain a set of items that are stylistically compatible with each other and compatible with the given scene. 
%The task is named as scene-based Complementary Item Retrieval. 
%To achieve this, we propose a two-stage self-supervise learning framework. 
%During the first stage, we learn the visual features that are necessary to do both similarity retrieval and compatibility retrieval. 
%At the second stage, we propose a Flexible Bidirectional Transformer for compatibility contrastive learning. 

\subsection{Conditional Compatibility Auto Reasoning}
% \TODO{connect the between 3.1 and 3.2.2}, [already exist in the end of this subsection]
%\JL{the sentence below doesn't make sense. also, keep the notation consistent.}
We formulate the problem of generating a set of objects $\mathcal{X}=\{x_i\}_{i=0}^L$ conditioned on the scene image $\mathcal{I}$ and a specified set of categories as how to compute likelihood (Eq.~\ref{eq:1}) of creating the object set $\mathcal{X}$ given the scene image $\mathcal{I}$, objects in the scene $\mathcal{O}$, and set of categories $\mathcal{C}$. We model the probability of generating the unordered set $\mathcal{X}$ as the sum of generating the set in any permutation $\hat{\mathcal{X}}$:
\vspace{-2mm}
\begin{equation}
%\resizebox{.9 \columnwidth}{!} {
p(\mathcal{X}_i|\mathcal{I},\mathcal{O},\mathcal{C}) =\sum_{\hat{\mathcal{X}}\in\Phi(\mathcal{X}_i)}p(x_i|x_0, \dots,x_{i-1}, \mathcal{I}, \mathcal{O},\mathcal{C}),i\leq L
%}
\label{eq:1}
\end{equation}
where $\Phi(\mathcal{X})$ includes all the permutation of the target object set $\mathcal{X}$ given all the permutation of the categories $\mathcal{C}$, and $L$ is the maximum number of items to compose a set.
%Note the items in the target set $\mathcal{X}$ and the objects in the source scene $\mathcal{O}$ are from two image domain.
%Here the domain means the distance $\mathcal{D}$ of the distribution of the pixels of object $\mathcal{X}$ and $\mathcal{O}$ are larger than a threshold $\theta$.
For each permutation of $\mathcal{X}$, the set generation becomes a sequence generation problem.
We model the sequence generation as an auto-regressive process.
In the auto-regressive process, the next item in the set is generated conditioned on the prior items.
This auto-regressive process statistically formulated as the multiplication of the probabilities:
\vspace{-2mm}
\begin{equation}
%\resizebox{.9 \columnwidth}{!} 
%{
    p(x_i|x_0,\dots,x_{i-1}) = \prod_j^{j<(i-1)} p(x_{j}|x_0,\dots,x_{j-1}).\label{eq:2}
%}
\vspace{-2mm}
\end{equation}
%    p(\mathcal{X}_i|\mathcal{I},\mathcal{O},\mathcal{C}) &=\sum_{\hat{\mathcal{X}}\in\Phi(\mathcal{X}_i)}p(x_i|x_0, \dots,x_{i-1}, \mathcal{I}, \mathcal{O},\mathcal{C}) \label{eq:1} \\
    %\mathcal{X}_i &= \{x_0, x_1, \dots, x_i\}, i \leq M \nonumber, x_i \in \mathcal{D}_B\\
    %C(\mathcal{X}_i) &= \mathcal{C}_i = \{c_0, c_1, \dots, c_i\}, i \leq M \nonumber\\
    %\mathcal{F}(\mathcal{D}_A,\mathcal{D}_B) &\geq \theta\nonumber \\
%    p(x_i|x_0,\dots,x_{i-1}) &= \prod_j^{j<(i-1)} p(x_{j}|x_0,\dots,x_{j-1})\label{eq:2}
%\end{align}
%We model the set compatibility as for any item $x_j\in\mathcal{X}_i$ in any order, the generated set $\mathcal{X}_i$ is self-compatible and compatible with the given scene $\mathcal{I}$.
To learn to conditionally generate the best set of objects, our model learns to maximize the log likelihood of the probability, $p(\mathcal{X}|\mathcal{I},\mathcal{O},\mathcal{C})$,
%\JL{you should give more detailed intuition first before dumping equations}
\vspace{-2mm}
\begin{equation}
%\resizebox{.9 \columnwidth}{!} 
%{
    \log p(\mathcal{X}_i|\mathcal{I},\mathcal{O},\mathcal{C}) =\sum_{\hat{\mathcal{X}}\in\Phi(\mathcal{X}_i)}\sum_j^{j<i}\log p(x_i|x_{<j}, \mathcal{I}, \mathcal{O},\mathcal{C}) 
 %   }
    \label{eq:3}
\end{equation}
%    \log p(\mathcal{X}_i|\mathcal{I},\mathcal{O},\mathcal{C}) &=\sum_{\hat{\mathcal{X}}\in\Phi(\mathcal{X}_i)}(\sum_j^{j<i}\log p(x_i|x_{<j}, \mathcal{I}, \mathcal{O},\mathcal{C})) \label{eq:3}\\
%    p(x_i|x_{j}, \mathcal{I}, \mathcal{O},\mathcal{C}) &= p(c_i|x_{j}, \mathcal{I}, \mathcal{O},\mathcal{C})p(\hat{x}_i|x_{j}, \mathcal{I}, \mathcal{O},\mathcal{C})
%\end{align}
To approximate the log likelihood Eq.~\ref{eq:3}, we propose a two-stage learning framework.

\subsection{Compatibility Learning Framework}
%In this section, we explain in details our proposed two-stage learning framework. 
%
%
%During the first stage, we learn a feature extractor that can project the object pixel information to an embedding that cluster objects that are similar and can represent features for compatibility at the same time. 
%And in the second stage, we propose a Flexible Bidirectional Transformer for compatibility reasoning. 
%\JL{in the second stage?}
%
\noindent\textbf{Visual Similarity Learning}
\begin{figure}[h]
\centering
\includegraphics[width=0.9\columnwidth]{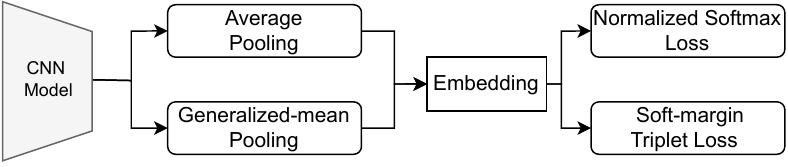}
\caption{\textbf{VSIM: Visual Similarity Model.} %We apply a standard ResNet50 followed by a combination of average pooling and generalized mean pooling global descriptors. We average the two pooling global descriptors as the final 2048-dimensional image embedding. This model is trained with Normalized Softmax Loss and Soft-margin Triplet Loss for retrieval task.
}
\label{fig:vsim}
\vspace*{-0.5em}
\end{figure}
%\subsubsection{First Stage: Visual Similarity Learning}
To relieve the complexity of learning both the visual complementarity and similarity directly from pixel domain, we propose to separate them into two stages. 
In the first stage, our model focuses on the visual similarity learning. 
%in the first stage, we found the feature embedding learned via visual similarity target is the best 
As shown in Figure~\ref{fig:vsim}, we apply a CNN-based (ResNet50) visual similarity model~\cite{jun2019combination} with normalized softmax loss~\cite{zhai2018classification} and soft-margin triplet loss~\cite{hermans2017defense} (refer to Supplementary for more details). 
With this model, we project the scene image $\mathcal{I}$, objects in the image $\mathcal{O}$, and the item images in retrieval pool $\mathcal{X}$ onto this embedding:
%\vspace*{-0.5em}
\begin{equation}
    \{\mathbf{I},\mathbf{O},\mathbf{X}\}=g(\{\mathcal{I}, \mathcal{O},\mathcal{X}\}), \mathcal{I}\in\mathbb{R}^3, \mathcal{O}\in\mathbb{R}^3, \mathcal{X}\in\mathbb{R}^3
\end{equation}
where $g$ is our visual similarity model.
%With the projected representation of the scene image $\mathbf{I}$, objects in the scene $\mathbf{O}$ and objects in the target domain $\mathbf{X}$, our second stage model doesn't need to start from the raw pixel space.
This projection helps our second stage model to converge faster, similar in spirit to sequential optimization.
%and learn the best representation for visual compatibility.
We also show empirically (refer to Sec. Similarity Learning Results for details) that the visual similarity embedding is best suited for learning cross-domain visual compatibility.
%\vspace{-0.5cm}

\noindent\textbf{Complementarity Reasoning with Flexible Bidirectional Transformer}
\begin{figure}[h!]
\centering
\includegraphics[width=1.0\columnwidth]{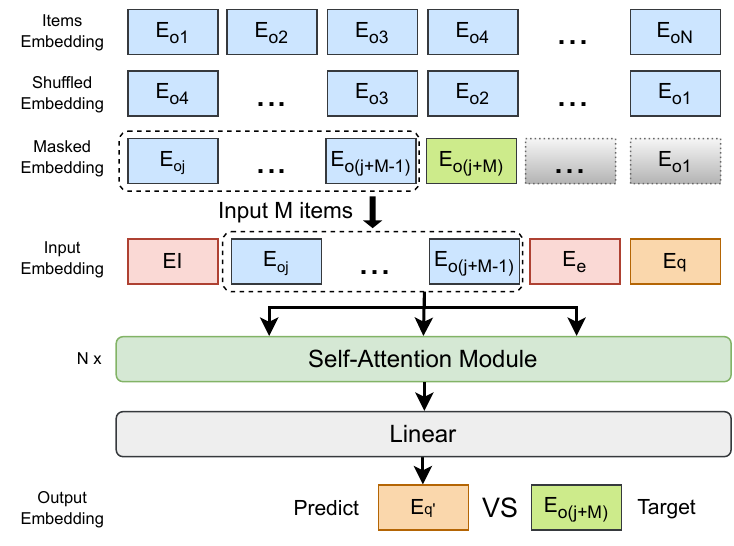}
\vspace{-1.5em}
\caption{\textbf{FBT: Flexible Bidirectional Transformer.} We randomly sample $M\in[0,N]$ items from the total item number $N$ of items (in a scene) as input set, and the $(M+1)_{th}$ item not in the input set as output target. We put the scene embedding at the beginning of input set, and take the scene embedding as the start token $EI$. We set a zero vector as the end token $E_{e}$. %The output will be the end token when the input set contains all the items in the scene. 
}
\vspace{-3mm}
\label{fig:fbt_overview}
\end{figure}
%\subsubsection{Second Stage: Compatibility Reasoning with Flexible Bidirectional Transformer}
%\TODO{refer the figure to supp, intuition}
At the second stage, we propose a new Flexible Bidirectional Transformer (FBT) (see Figure ~\ref{fig:fbt_overview} for \textit{conditional cross-domain unordered set generation}).
We choose Transformer model~\cite{vaswani2017attention} as the core architecture to learn the inter-object compatibility.
%We innovate upon the vanilla Transformer model~\cite{vaswani2017attention} to self-supervisely learn the visual compatibility between a set of unordered items. 
The vanilla Transformer model~\cite{vaswani2017attention}, as it is originally proposed for modeling ordered sequence structured data, such as languages and images, is insufficient for our task. 
%We need to innovate it for \textit{conditional cross-domain unordered set generation}. 
%Thus, we propose a new Flexible Bidirectional Transformer (FBT) (Figure ~\ref{fig:fbt_overview}, refer to Supplementary Section 3 for the detailed illustration) to approximate the likelihood (Eq.~\ref{eq:1}) of predicting the next item $x_i$ given the scene $\mathcal{I}$, objects in the scene $\mathcal{O}$ with or without specific categories $\mathcal{C}$. 

We introduce: 
%with a trainable token, 
(1) random shuffling together with random length sequence masking for set generation;
(2) category prediction arm to better model the category distribution for a set of objects;
and (3) visual embedding prediction arm for visual compatibility modeling. 
During inference, our FBT model generates an unordered set auto-regressively (Eq.~\ref{eq:2} and shown in the green part of Figure~\ref{fig:model_overview}).
%We also demonstrate this auto-regressive generation in the green portion in Figure~\ref{fig:model_overview}). 
%Refer to Figure 1 in Supplementary Section 3 for the details.
%With the random shuffling and random length sequence masking 
Inspired from the CLS token proposed in the Vision Transformer (ViT)~\cite{dosovitskiy2020image}, we also use a trainable variable denoted as $\mathbf{q}$ to extract inter-token relation,
% %The corresponding output of the trainable variable denoted as $\mathbf{q}'$ is,
% \begin{align}
%     \mathbf{q'} &= e(\phi(\mathbf{E}\mathbf{X});\mathbf{E}\mathbf{I},\mathbf{E}\mathbf{O},\mathbf{E}\mathbf{q})\\
%     &=\text{MLP}(\text{MSA}(\phi(\mathbf{E}\mathbf{X});\mathbf{E}\mathbf{I},\mathbf{E}\mathbf{O}))\\
%     \phi(\mathbf{E}\mathbf{X})&=[\mathbf{E}\mathbf{x}_1, \mathbf{E}\mathbf{x}_2,\dots,\mathbf{E}\mathbf{x}_N,\text{MASK}]\nonumber\\ \mathbf{E}\mathbf{O}&=[\mathbf{E}\mathbf{o}_1,\mathbf{E}\mathbf{o}_1,\dots,\mathbf{E}\mathbf{o}_M]\nonumber
% \end{align}
\begin{align}
\begin{split}
%\vspace{-4mm}
\mathbf{q'} &= e(\mathbf{EI},\Phi(\mathbf{E}\mathbf{O}); \mathbf{E}\mathbf{q}) \\ &=\text{MLP}(\text{MSA}(\mathbf{EI},\Phi(\mathbf{E}\mathbf{O}), \mathbf{E}\mathbf{e},\mathbf{E}\mathbf{q}))\\
%\Phi(\mathbf{E}\mathbf{X}) &= \Phi(\mathbf{E}\mathbf{I},\mathbf{E}\mathbf{O}) = [\mathbf{E}\mathbf{x}_I,\mathbf{E}\mathbf{x}_O, \mathbf{E}\mathbf{x}_e, \text{MASK}]\\
%&=[\mathbf{E}\mathbf{x}_1, \mathbf{E}\mathbf{x}_2,\dots,\mathbf{E}\mathbf{x}_N,\text{MASK}] \\
\mathbf{E}\mathbf{O} &=[\mathbf{E}\mathbf{o}_1,\mathbf{E}\mathbf{o}_2,\dots,\mathbf{E}\mathbf{o}_{M}, \text{MASK}],%\nonumber
\end{split}
%\vspace{-3mm}
\end{align}
\noindent
where $\mathbf{q}'$ denotes the corresponding output of the trainable input token $\mathbf{q}$, $\mathbf{e}$ is the end token, $\Phi$ is the masking operation, $e()$ represents the Transformer encoder with MSAs (the Multi-headed Self-Attention layers), MLPs (Multi-Layer Perception), $\mathbf{E}$ is the linear projection and $M$ is the unmasked sequence length.
%, $\mathbf{x}_I$ is the scene embedding and $\mathbf{x}_e$ is the end token, $N$ is the number of input.  And $\mathbf{o}_i$ is one of the embedded object in the scene $\mathbf{O}$, $M$ is the number of object in the scene. 
The output $\mathbf{q}'$ is then be used for predicting both the category $\mathbf{c}_{M+1}$ (Eq.~\ref{eq:7}) and the visual embedding of the next item $\mathbf{x}_{M+1}$ (Eq.~\ref{eq:8}).
\begin{align}
%\vspace{-4mm}
    \mathbf{\hat{c}} &=\text{MLP}(\mathbf{q}') \label{eq:7}\\
    \mathbf{\hat{x}_{M+1}}&=\text{MLP}[\mathbf{q}',\mathbf{\hat{c}}]\label{eq:8}
%    \vspace{-4mm}
\end{align}
The output category embedding $\mathbf{\hat{c}}$ is supervised via the Cross-Entropy loss. 
%\paragraph{Cross-Entropy Loss:} To predict the class labels of the output tokens, we use the the softmax function to convert the outputs into probabilities of target classes. We minimized the cross-entropy loss for the class prediction.
And the visual feature embedding $\mathbf{\hat{x}}$ 
%for the compatible image retrieval 
is supervised using a triplet loss~\cite{yang2019xlnet}. 
To form a triplet, the anchor is the predicted embedding $\mathbf{\hat{x}_{M+1}}$ with the target item's embedding $\mathbf{x_{M+1}}$ as the positive and randomly selected same category object's embedding as the negative. 
%The positive of the triplet is the target item's feature embedding $\mathbf{x_{M+1}}$.
%generated by the first stage visual similarity model. 
%As for the negative(s), we randomly select a same category object from the target domain.
%and generate their feature embedding through the first stage visual similarity model.
%\paragraph{Differential Entropy Loss:} 
One challenge in the feature learning is the space collapsing, where points in the embedding space get too close.
This space collapse can lower the representation capacity. 
To avoid such an issue, we apply a differential entropy regularizer~\cite{girdhar2019video} to maximize the distance between each point and its closest neighbors in the embedding space. 
The regularizer is defined as follows:
\begin{equation}
\vspace{-2mm}
\text{Reg}=-\frac{1}{N}\sum_{i}^{N}log(D_{min_{i\ne j}}(z_i, z_j)),
\end{equation}
where we divide the L2 distance between sample i, j by 4 as $D_{i,j}$ to make the distance between 0 and 1. $z$ is the input feature maps. %We obtained $0.9\%$ improvement from this regularizer.

\section{Experiments}
%To illustrate the effectiveness of our method, we evaluate the proposed ICAR on dataset DeepRooms~\cite{gadde2021detail}, STL-Home~\cite{kang2019complete}, STL-Fashion~\cite{kang2019complete}, and Exact Street2Shop~\cite{hadi2015buy} using the FITB and our proposed SFID metrics.
%For DeepRooms, it consists of 1.3 million images, including lifestyle image, cropped furniture images, and white background product images. For STL, it has home dataset and fashion dataset, consisting of 151 k images. Especially, STL provides data pairs that products have a similar style to the observed items in the scene. For Exact Street2Shop, it contains 15 k images for fashion related task. 

\subsection{Setup}
\label{sec: experimental_setup}
\noindent\textbf{Benchmark Datasets: } In the following experiments, we evaluate our proposed ICAR using four datasets. 
DeepRooms~\cite{gadde2021detail} is a large-scale (1.4 million), high-quality human annotated object detection dataset covering a total of 81 fine-grained furniture and home product categories with 210K room-scene images.
%, we use 13k scene image, 626k cropped furniture images, and 719k wight background product images for our experiments. Class labels and locations of furniture and decor products are manually annotated on the images with each product also linked to a plain-background product image.  All the On average, each scene contains 5 to 6 furniture.
STL-Home~\cite{kang2019complete} includes 24,022 interior home design images and 41,306 home decor items, which can generate 93,274 scene-product pairs. 
%And this dataset has bounding boxes for products, corresponding plain-background product images, and product category information, all of which are labeled by human workers. 
STL-Fashion~\cite{kang2019complete} contains 72,189 fashion-product pairs from its 47,739 fashion images and 38,111 product images. 
%Similar to STL-Home, this collection includes bounding boxes for products, related plain-background product photos, and information on product categorization.
And Exact Street2Shop~\cite{hadi2015buy} 
%introduces the first human-labeled dataset for the street2shop task. This dataset 
provides fashion-product 10,608 pairs from its  10,482 fashion images and 5,238 product images with the bounding box of products in scene images.

% may update the scene set pair number --xjw
%The Visual Scene-based Complementary Item Retrieval (CIR) application allows users to quickly search for and purchase a product (or one that is similar) by taking a picture of it when they see it anywhere. The fundamental issue here is that products in online shopping photographs are often in a canonical position, on a plain background, appropriately lighted, etc., in contrast to those in real-world scenes, where products can be found in a variety of poses, background and settings. Therefore, we construct all datasets in the following format:
%\begin{itemize}
%\item Every scene matches with a set of cropped furniture and the corresponding plain background product images. Since most of the scene images come from designers, so we assume the cropped furniture from one scene are stylistically compatible.
%\item We prepare a product bank for positive and negative sampling, and the product bank is organized by category.
%\item All images will go though the same backbone to extract feature embeddings.
%\end{itemize}
%This kind of data format allows us explore the style matched complementary item retrieval application in a self-supervised manner.
\noindent\textbf{Implementation:}
All the models in our experiments are trained using AdamW~\cite{loshchilov2017decoupled} with cosine learning scheduler from an initial learning rate of $2e-4$ to 0. 
Models are trained for 500 epochs with a batch size of 256. 
%Before feeding them to the transformer encoder, we transfer the 2048-dimensional image embedding from a visual similarity backbone model to 256-dimensional sequence embeddings. 
We choose 1 negative sample in the triplet loss. 
And we use 1.0, 1.0, and 0.05 as the weights for cross-entropy loss, triplet loss, and regularizer loss respectively.

\begin{figure*}[t]
%\vspace{-4mm}
\centering
\includegraphics[width=1.0\textwidth]{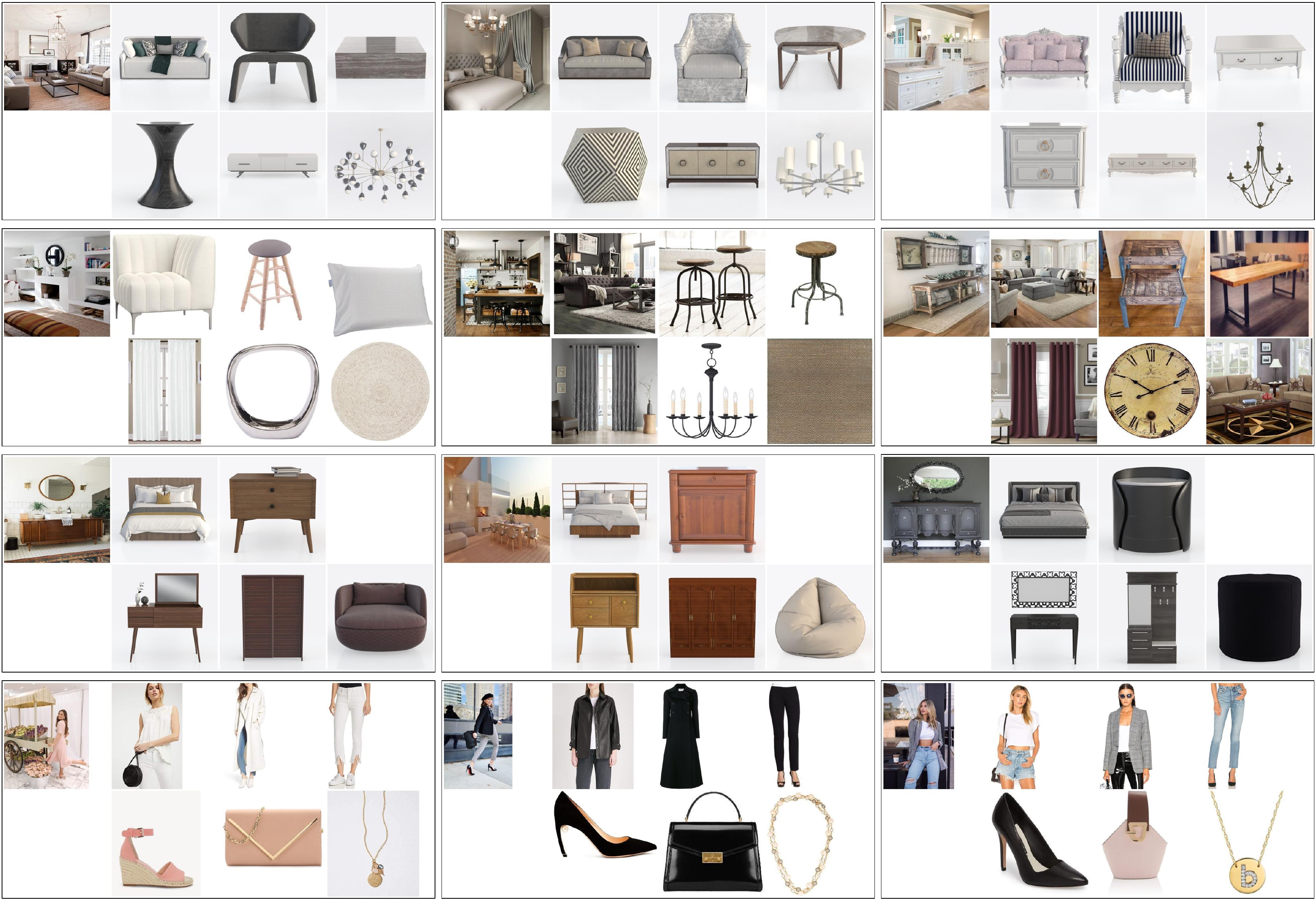}
\vspace{-5mm}
\caption{\textbf{Scene-aware Cross-Domain CIR Qualitative Results.} 
We show qualitatively our model is capable of retrieving stylistically compatible items from both {\em seen} (Row 2 and 4) and {\em unseen} domains (Row 1 and 3), given a home (Row 1-3) or fashion (Row 4) scene image.
Column 1, 5, 9 are the input scene images.
See supplementary materials for more examples.
%from left to right and from top to bottom, the first 3 sets are living room recommendation, the second 3 sets are STL-Home recommendation, the third 3 sets are bedroom recommendation, the last 3 sets are fashion recommendation. Given an inspirational scene image (living room, STL-home, and bedroom are sampled from STL-home, and fashion is sampled from STL-fashion as shown in column 1, 5, 9) with a pool of products (living room and bedroom are from 3D-FRONT, STL-home and STL-fashion are from STL). Our model auto-regressively retrieves a set of stylistically compatible items (column 2,3,4 6,7,8,10,11,12 in the picture). Please refer supplementary for more examples.
}
\label{fig:visual}
\vspace{-1.2em}
\end{figure*}

\subsection{Evaluation Metrics}
\label{sec:evaluation_metrics}
To evaluate our proposed model's performance for scene-aware cross-domain CIR, we benchmark our method using previously proposed Fill-In-The-Blank (FITB)~\cite{han2017learning} and our newly proposed SFID metrics.
FITB accuracy is measured via counting percentage of times the model correctly picked the ground-truth object from the candidate pool.
Following \cite{kang2019complete}, we set the number of candidate to 2 for STL and Street2Shop and 3 for DeepRooms. 
For fair comparison, we apply the same method in setting up the FITB candidate pool across all the experiments.
While FITB estimates the model's ability to retrieve the best object, it 
fails to capture the compatibility among a set of items. 
To compensate, we propose a new distribution distance-based metric: {\bf Style FID (SFID)}. 
It is difficult to define visual style, as it covers various  aspects, including color, texture, shape and so on. 
Instead, we use the {\em visual similarity distribution} between the generated set and the designed sets of items as the measurement of stylistic compatibility.
Similar to the FID~\cite{heusel2017gans}, we apply the Fr\'echet distance $\mathcal{F}$ for distribution distance measurement.
Instead of directly estimating the pixel value distribution, we apply a feature extractor that can project the pixel values onto an embedding.
With the feature extractor, the {\em computed distribution} can focus on style related features, including colors, edges, textures, patterns, and shapes. We define
%For two sets of products, we paste two image sets into two picture separately, and measure the Style-FID score,
\begin{equation}
\text{SFID Score} = \mathcal{F}(f(\mathcal{X}), f(\mathcal{Y})),
\end{equation}
\begin{equation}
\mathcal{F}(\mathbf{X}, \mathbf{Y})=\mid\mu_X-\mu_Y\mid + tr(\sigma_X+\sigma_Y-2\sqrt{\sigma_X*\sigma_Y }),\nonumber
\end{equation}
%\TODO{add back the Frechet inception score}
where $\mathcal{X}$ is the generated set of objects, $\mathcal{Y}$ is a set of well designed (ground-truth) objects, function $f$ is a feature extractor, $\mu$ and $\sigma$ are the mean and variance.
Please refer supplementary for more details about SFID. 
\subsection{Compatibility Learning Results}
\begin{table}
\centering
\resizebox{0.95\columnwidth}{!}{
\begin{tabular}{c c c }
\toprule
Method & FITB $\uparrow$ & Condition  \\
\midrule 
CSA-Net w. triplet loss  & $60.0$ & / \\
CSA-Net w. outfit loss  & $57.4$ & / \\
Visual Similarity Learning  & $62.0$ & / \\
OutfitTransformer  & $77.6$ & / \\
\midrule 
\textbf{ICAR (Ours)}  & \textbf{83.9} & Predict  category \\
\textbf{ICAR (Ours)}  & \textbf{87.1} & Given category \\
\bottomrule 
\end{tabular}
}
%\vspace{1mm}
\caption{{\bf FITB Results on DeepRooms~\cite{gadde2021detail}.} Our approach improved FITB accuracy by {\bf 9.5\%} over Visual Similarity Learning, CSA-Net~\cite{lin2020fashion}, and OutfitTransformer~\cite{sarkar2022outfittransformer}. }
\label{tab:style_home}
\vspace{-1em}
\end{table}

% \begin{table}
% \centering
% \resizebox{0.95\columnwidth}{!}{
% \begin{tabular}{c c c }
% \toprule
% Method & FITB $\uparrow$ & SFID $\downarrow$   \\
% \midrule 
% CSA-Net w. triplet loss \cite{lin2020fashion} & $60.0$ & / \\
% CSA-Net w. outfit loss  \cite{lin2020fashion} & $57.4$ & / \\
% %CGD \cite{qiuying2019} & $62.0$ & / \\
% CLTR (Sec. 1 Supplementary)   & $77.6$ & $48.0$ \\
% \midrule 
% %\textbf{ICAR (Ours)}  & \textbf{83.9} & Predict the target category \\
% \textbf{ICAR (Ours)}  & \textbf{87.1} & \textbf{36.8} \\%Given the target category \\
% \bottomrule 
% \end{tabular}
% }
% \caption{{\bf Quantitative Results on DeepRooms~\cite{gadde2021detail} Dataset.} Our approach improved the FITB accuracy by 9.5\% and SFID score by 12.8\%.}
% \label{tab:style_home}
% \end{table}
\subsubsection{Quantitative Results}
Here we present quantitative evaluation results on four different scales for benchmark datasets.
We first evaluate our algorithms on the largest DeepHomes~\cite{gadde2021detail} dataset. 
As shown in Table~\ref{tab:style_home} and ~\ref{tab:sfid_stl}, ICAR improves performance over the state-of-the-art by {\bf 9.5\%} for FITB and 11.2 (about {\bf 23.3\%}) for SFID score.
To further test the effectiveness of our category embedding, we compare the FITB results when specify or not the category during inference time.
If category is not specified, the FITB score drops from $87.1\%$ to $83.9\%$.
%we conduct our experiments in different settings, one is to predict the product category and the other is to specify the category. 
%Based on our results, we observe that specifying the category outperforms predicting the product category. 
%For predicting the product category, our ICAR improves performance over the state-of-the-art by 6.3\%. 
%For specifying the category, . As shown in Table~\ref{tab:style_home_sfid}, for our proposed SFID, ICAR improves performance over the state-of-the-art by {\bf 12.8\%}. 
%All those results illustrate the effectiveness of our methods.

%\subsection{Results on STL}
\begin{table}[!h]
%\vspace{-3mm}
\centering
\resizebox{1.0\columnwidth}{!}{
\begin{tabular}{c c c c}
\toprule
Method & STL-Fashion $\uparrow$ & STL-Home $\uparrow$ & S2S $\uparrow$   \\
\midrule 
IBR  & 58.5 & 57.0 & 56.5 \\
Siamese Nets   & 67.1  & 72.4 & 63.0 \\
BPR-DAE  & 61.1 & 64.2 & 59.3 \\
Complete the Look  & 70.0 & 75.0 & 63.1 \\
OutfitTransformer & 65.0 & 77.0 & 67.2 \\
\midrule 
\textbf{ICAR (Ours)}  & \textbf{75.3} & \textbf{86.6}&\textbf{71.4} \\
\bottomrule 
\end{tabular}
}
\vspace{-0.2em}
\caption{{\bf FITB Results on STL and Street2Shop (S2S).} Our approach can improve FITB accuracy by {\bf 4.2\%--9.6\%} over IBR~\cite{mcauley2015image}, Siamese Nets ~\cite{veit2015learning}, BPR-DAE~\cite{song2017neurostylist}, Complete the Look ~\cite{kang2019complete}, \& OutfitTransformer~\cite{sarkar2022outfittransformer}.}
\label{tab:stl}
\vspace{-0.8em}
\end{table}

We then evaluate our method on more existing datasets: STL-Home~\cite{kang2019complete}, STL-Fashion~\cite{kang2019complete} and Street2Shop~\cite{hadi2015buy}. 
For these three datasets, the human annotators also label products that have a similar style to the observed product and are compatible with the scene. 
This study suggests that the scene images and the products may not have one-to-one correspondence, thereby increasing the difficulty of style matching. 
We show in Table~\ref{tab:stl} and Table~\ref{tab:sfid_stl}, that ICAR outperforms SOTA by 5.3\% on STL-Fashion and 9.6\% on STL-Home in FITB metric, and outperforms SOTA by 3.4 (22.3\%) on STL-Fashion and 2.9 (31.8\%) on STL-Home in terms of SFID metric.

\begin{table}[!h]
\centering
\resizebox{1.0\columnwidth}{!}{
\begin{tabular}{c c c c c}
\toprule
Method & DeepRooms $\downarrow$ & STL-F $\downarrow$ & STL-H $\downarrow$ & S2S $\downarrow$  \\
\midrule 
GT VS Neg & 49.0 & 15.8 & 10.2 & 9.1 \\
OutfitTransformer  &48.0 & 15.2 & 9.1 & 8.7\\
\midrule 
\textbf{ICAR (Ours)}  &\textbf{36.8} & \textbf{11.8} & \textbf{6.2} & \textbf{7.1} \\
\bottomrule 
\end{tabular}
}
\vspace{-2mm}
\caption{{\bf SFID Results on DeepRooms, STL(F: fashion, H: home) and Street2Shop(S2S).} Our approach can improve SFID accuracy by 11.2 ({\bf 23.3\%}, DeepRooms) and 2.9 ({\bf 31.8\%}, STL-Home) on furniture and by 3.4 ({\bf 22.3\%}, STL-F) and 1.6 ({\bf 18.4\%}, S2S) on fashion images, respectively, over OutfitTransformer~\cite{sarkar2022outfittransformer}.}
\label{tab:sfid_stl}
\vspace{-4mm}
\end{table}
\begin{figure*}[t]
%\vspace{-2mm}
\centering
\includegraphics[width=1.0\textwidth]{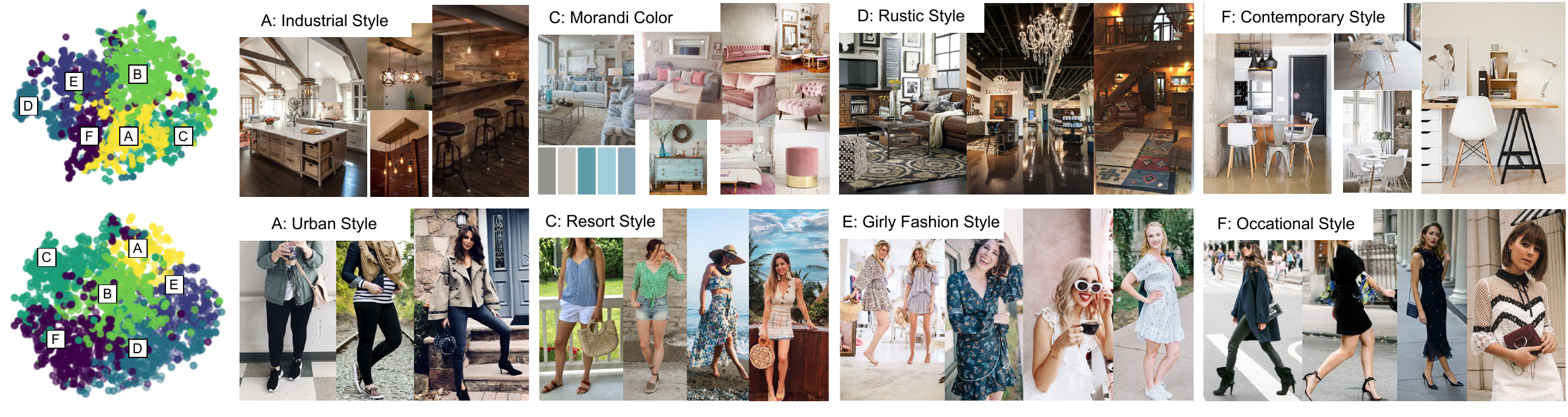}
\vspace{-8mm}
\caption{\textbf{Learned Scene Image Embedding Clustering Results.} To validate the style implicitly learned through our network, First column is the t-SNE on the randomly sampled 2k STL-home and STL-fashion test in split scene images (Columns 2-5). % are the identified fashion styles and interior design styles in those clusters. 
}
\label{fig:tsne}
\vspace{-2mm}
\end{figure*}
\subsubsection{Qualitative Results}
Next we present qualitative results on scene-based cross-domain CIR results.
%We test our ICAR model on two scenarios: retrieve from a domain our model has seen and from an unseen one.
%
We sample scene images from the STL-home and STL-fashion dataset test-split.
Then we specify the item categories. 
%We test our model using three sets of item categories: 1) sofa, chair, coffee table, end table, tv-stand, ceiling light; 2) bed, nightstand, vanity, dresser, chair; 3) top, bottom, out ware, shoes, bag, jewllery. 
As shown in Figure~\ref{fig:visual}, from left to right and from top to bottom, our model generates diverse style recommendations, including but not only Contemporary (e.g. 1st, 2nd, 4th and 9th set), Classic (e.g. 3rd, 7th and 8th set), Industrial (e.g. 5th set), Rustic (e.g. 6th set). Refer to supplementary materials for more examples. 
%\sout{Furthermore, following \cite{kang2019complete}, we also conduct the binary FITB (1 negative candidate) experiments, as shown in Figure~\ref{fig:fitb}, our model chooses the more stylistically compatible item in the green box.}

\subsubsection{Sequence Masking Validation}
\begin{table}
\centering
\resizebox{0.95\columnwidth}{!}{
\begin{tabular}{c c c }
\toprule
Method & FITB $\uparrow$ & Condition   \\
\midrule 
ICAR - fixed length masking & 82.2 & Predict category \\
ICAR - random length masking & 83.9 & Predict category \\
ICAR - fixed length masking & 84.8 & Given category \\
ICAR - random length masking & \textbf{87.1} & Given category \\
\bottomrule 
\end{tabular}
}
\vspace{-2mm}
\caption{{\bf Masking Method Comparison.} Our random length masking outperforms fixed length masking.}
\label{tab:sample}
\vspace{-5mm}
\end{table}
%\TODO{Need clean up}
%For visual scene-aware Complementary Item Retrieval (CIR) application, you never know what scene image users will input. 
%Therefore, we proposed a flexible bidirectional transformer to train a deep bidirectional representation. 
In our FBT model, we propose a random shuffle and random length masking technique for unordered set generation.
Here we ablate our proposed random length masking technique.
In this experiment, we compare the results when we apply random length masking vs. fixed length masking to the input sequence.
%For every scene image, we first sample a random length of product embeddings in this scene, and then random pick one embedding as output and the others are the input set. 
%Both the input length and the product embeddings are random during training. 
As shown in Table~\ref{tab:sample}, we show that our random length masking approach outperforms fixed length masking for unordered set generation with or without target category been given.

\subsubsection{Visual Compatibility Embedding Validation}
%\TODO{Need clean up}
To validate our model's ability in learning the visual style implicitly, we perform t-SNE analysis on the learned embedding of the scene images sampled from the STL-home and STL-fashion datasets. 
In Figure~\ref{fig:tsne}, we show the clustering results on the left and the scene images from the clusters in column 2-5.
The cluster labels are computed using the k-means ($K=6$) method.
We find some typical interior design styles such as the industrial style, rustic style and contemporary style in the clusters.
More interestingly, we also find that our model  pays attention to color. 
For example, in the C home scene-image cluster, our model learns to extract the Morandi type of color scheme.
Given that our model is trained in a self-supervised way, we observe some mix in style within some of the clusters.
%visualize the learned embedding, 
%as shown in Figure, our learned embedding indeed contains style information. We use the learned embedding to do style clustering with k-means method, then we use TSNE to do the visualization. The STL-home scene images and STL-fashion scene images in the same cluster has similar style.

%of the linear projected scene images. 
%With k-means method, the STL-home scene images and STL-fashion scene images 

\subsubsection{Human Perception Validation}
\label{sec:user_study}
\begin{figure}
\centering
\includegraphics[width=1.0\columnwidth]{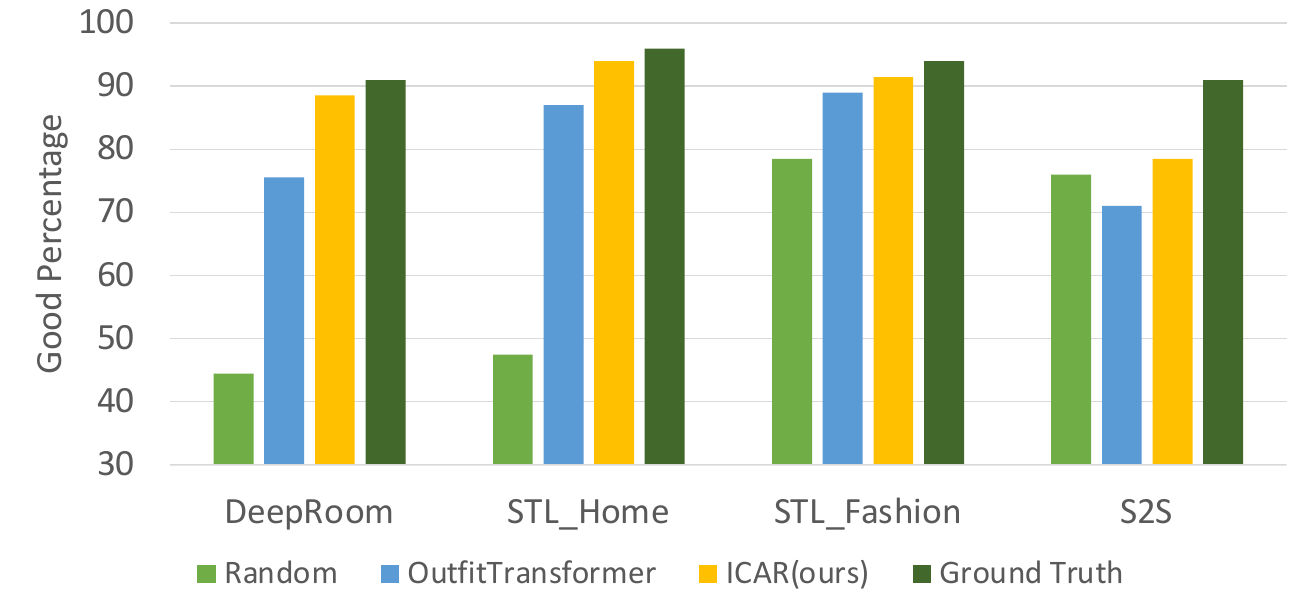}
\vspace{-5mm}
\caption{\textbf{Human Ratings on Different Datasets.} Our SFID score correlates better than SOTA with human judgement.
}
\vspace{-5mm}
\label{fig:userstudy}
\end{figure}
We conduct user studies by asking interior design experts to rate (Good: items are compatible, Neutral: one or two items are incompatible) 100 generated sets of furniture item images from the Random sample, OutfitTransformer\cite{sarkar2022outfittransformer}, our method (ICAR), and Ground Truth for different datasets. As shown in Figure~\ref{fig:userstudy}, ICAR outperforms all other methods, and those results are consistent with our SFID score (Table~\ref{tab:sfid_stl}). We further compute the Pearson correlation coefficient to measure the association between SFID and human rating score (normalized as Ground truth score/Method score). 
We obtain a 0.7 average Pearson correlation value ([-1, 1]).
This demonstrates that there is a strong positive association between our SFID score and human perception. 
%This demonstrates our measurement is congruous to human judgment.
%CLTR gets 29.75\% Good and 70.25\% Neutral. ICAR gets 53.50\% Good and 46.50\% Neutral.}

%In this section, we ablate the key components of our method to validate the effectiveness.  
%First, we  our assumption that focusing on visual similarity learning on the 
%Firstly, we are trying to explore what kind of backbone is suitable for scene-based CIR. 
%Then we will evaluate the flexible bidirectional transformer sampling method, loss functions, proposed metric SFID, T-SNE visualization. The ablation experiment is conducted on DeepRooms. The experimental setting on this dataset is the same as that in section~\ref{sec: experimental_setup}.
%\vspace{-2mm}

\subsection{Similarity Learning Results}
\label{sec: backbone}
Here we validate the effectiveness of focusing on visual similarity learning in the first stage.
%In most prior CIR methods, they use the Resnet pre-trained on ImageNet as the backbone. 
%To study the functionality of the proposed visual similarity model, 
We compare our method against four different types of learning target i.e. image classification, image reconstruction, image generation and image representation.
And for each type of learning target, we choose the state-of-the-art method (shown in the first column in Table~\ref{tab:backbone}) except the image representation. 
For image representation, %\TODO{add details here}
we train the model in a contrastive learning manner.
%in order to get better style feature.
All the models in this experiment are trained using the DeepRooms~\cite{gadde2021detail} dataset.
As shown in Table~\ref{tab:backbone}, our model performs the best when the first stage focus on the visual similarity learning.
%We go a further step to explore what kind of backbone works better on our scene-based CIR task. 
%As shown in Table~\ref{tab:backbone}, we have tried Classification, Reconstruction, Generation, Image Representation, and Retrieval tasks. 
%We report the best accuracy we can get and demonstrate that retrieval model is more suitable for scene-based CIR. 
\begin{table}
%\vspace{-3mm}
\centering
\resizebox{0.95\columnwidth}{!}{
\begin{tabular}{c c c }
\toprule
Method & FITB $\uparrow$ & Learning Target Type   \\
\midrule 
ICAR - VQGAN & $73.1$ & Reconstruction \\
ICAR - Swin & $80.5$ & Classification \\
ICAR - BEiT & $73.3$ &  Generation\\
ICAR - Contrastive Learning   & $75.0$ & Image Representation \\
ICAR - Visual Similarity  & $87.1$ & Retrieval \\
\bottomrule 
\end{tabular}
}
\vspace{-2mm}
\caption{{\bf Similarity Learning:} visual similarity learning is the most suitable for scene-based CIR. VQGAN \cite{esser2021taming}, Swin  \cite{liu2022swin}, BEiT \cite{bao2021beit}}
\label{tab:backbone}
\vspace{-1.2em}
\end{table}

% \subsubsection{Compatibility Model}
% \XW{
% We design two different transformers for compatibility modeling, one is Complementary Transformer (CLTR), and the other is Flexible Bidirectional Transformer (FBT). Compared with other existing methods,  CLTR is already a SOTA transformer-based framework for CIR task. Compared with ICAR, CLTR
% predicts a set of complementary items at once and is not category-aware. CLTR also learns to generate sets of complementary items without the need of annotations by professionals. Refer to section 1 of the supplementary for more details. We choose to use FBT in our framework given its flexibility and better performance (As shown in Table ~\ref{tab:style_home}, ~\ref{tab:stl} and ~\ref{tab:sfid_stl}).
% }
%\subsection{Ablation Study}

\section{Conclusion, Limitation, and Future Work}
%It is hard to abstract or describe the essence of a good home design or a fashion outfit (style). This subjectivity makes the visual compatibility even more challenging to model computationally. 
In this paper, we introduce a {\em compatibility learning framework} using a novel category-aware ``Flexible Bidirectional Transformer'' (FBT), based on {\em Visual similarity} and {\em complementarity} to effectively retrieve a set of stylistically compatible items across domains, given a scene-image query.
%Our model effectively learns both the similarity and complementarity between set of items. 
%Our model doesn’t require complementary labels. 
% Instead of modeling visual complementary as pairs or ordered sequence, we adopt {\em unordered set complementary}. 
This learning framework is also generalizable and can be extended to other types of conditional cross-domain CIR tasks. 

While our results show a promising direction, there is more to be explored. 
% input diversity \TODO{expand the input}
First, there are other informations, like text descriptors and video contents, to be used as compatibility signals. 
%\TODO{expand the style definition to broader}
Secondly, the metric to measure styles can be further broadened to encode regional preference and cultural influence,  interpretation of styles can be expanded to include more global and diverse perspectives.
%We conduct intensive ablation studies and experiments, our method achieves 9.5\% FITB score improvement and 11.2 SFID improvement compared with the SOTA methods, which illustrate the stable performance and the convincingness of our proposed ICAR. In the future, we will try to explore the possibility to combine visual similarity and compatibility learning.

\begin{small}
\bibliography{aaai24}
\end{small}

\newpage
\title{ICAR: Image-based Complementary Auto Reasoning - Supplementary Material}
\section{Supplementary Material}
\subsection{Implementation Details}
For our visual similarity learning, we apply a standard ResNet50 followed by a combination of average pooling and generalized mean pooling global descriptors. 
We average the two pooling global descriptors as the final 2048-dimensional image embedding. This model is trained with Normalized Softmax Loss and Soft-margin Triplet Loss for retrieval task.

In our complementarity learning, we employ 6 self-attention layers with 8 heads. 
The input and output token dimension is 256. 
We set the number of output query to 1 for training and 9 for testing. 
During the training phase, we randomly shuffle all the items in the scene and randomly mask out $[0,M], M \leq N$ (where N is the total number of items in the scene) items in the sequence as input.
%sample 0 to the total number of items in the scene as the input token sequence.
%Then the model needs to predict the one item by inputting all the other item in the masked out items. 
Then the model predicts the next $(M+1)th$ item in the sequence. 
We put the scene embedding at the beginning of input set as the start token. 
We set a zero vector as the end token. 
%The output will be the end token when the input set contains all the products in the scene. 
During testing phase, we use the auto-regressive manner to predict products and stop when the predicted output number reach 9 or the predicted output is the end token. For OutfitTransformer\cite{sarkar2022outfittransformer}, we use our Visual Similarity model as encoder and take it as the visual compatibility learning model. To keep the fair competition,  we replace the text input with the label input and keep all the training setting the same as in other experiments in this paper.

\begin{figure}[t]
\centering
\includegraphics[width=1.0\columnwidth]{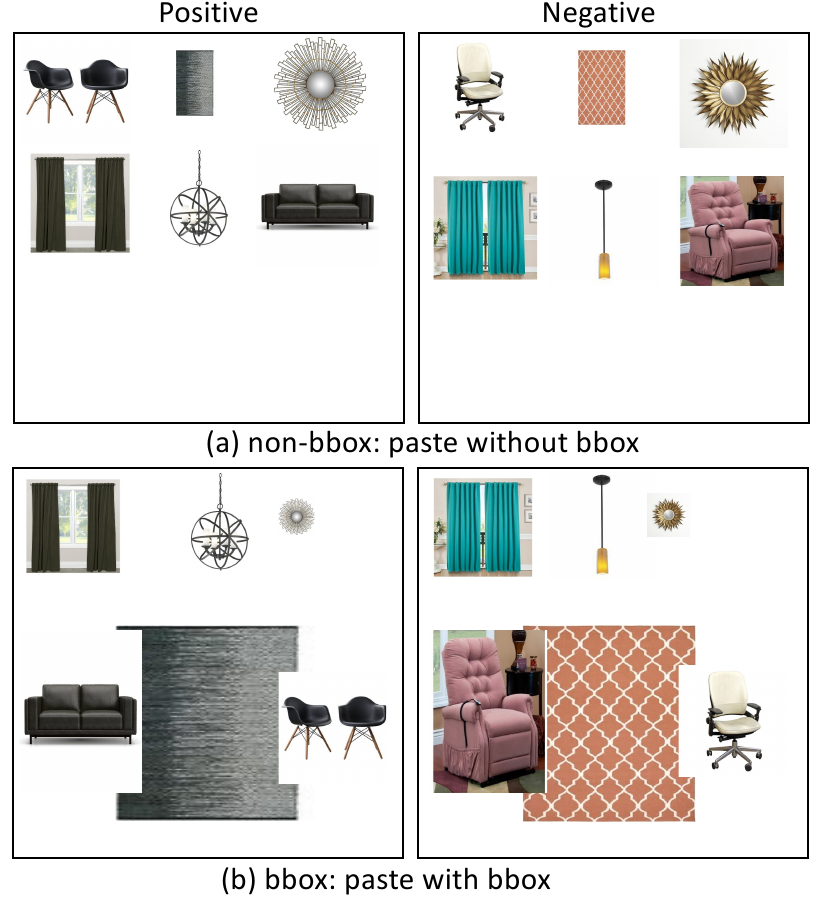}
\caption{\textbf{SFID Composed Images Comparison.} Here we show how we compose the set images. In (a), we randomly place the item images with fixed size (same height, aspect ratio reserved). In (b), we place the item images using the bbox of the item in the original scene images. After the study, we apply (a) in our SFID computation.
}
\label{fig:sfid}
\vspace{-3mm}
\end{figure}

\begin{figure*}
\centering
%\vspace*{-1em}
\includegraphics[width=1.0\textwidth]{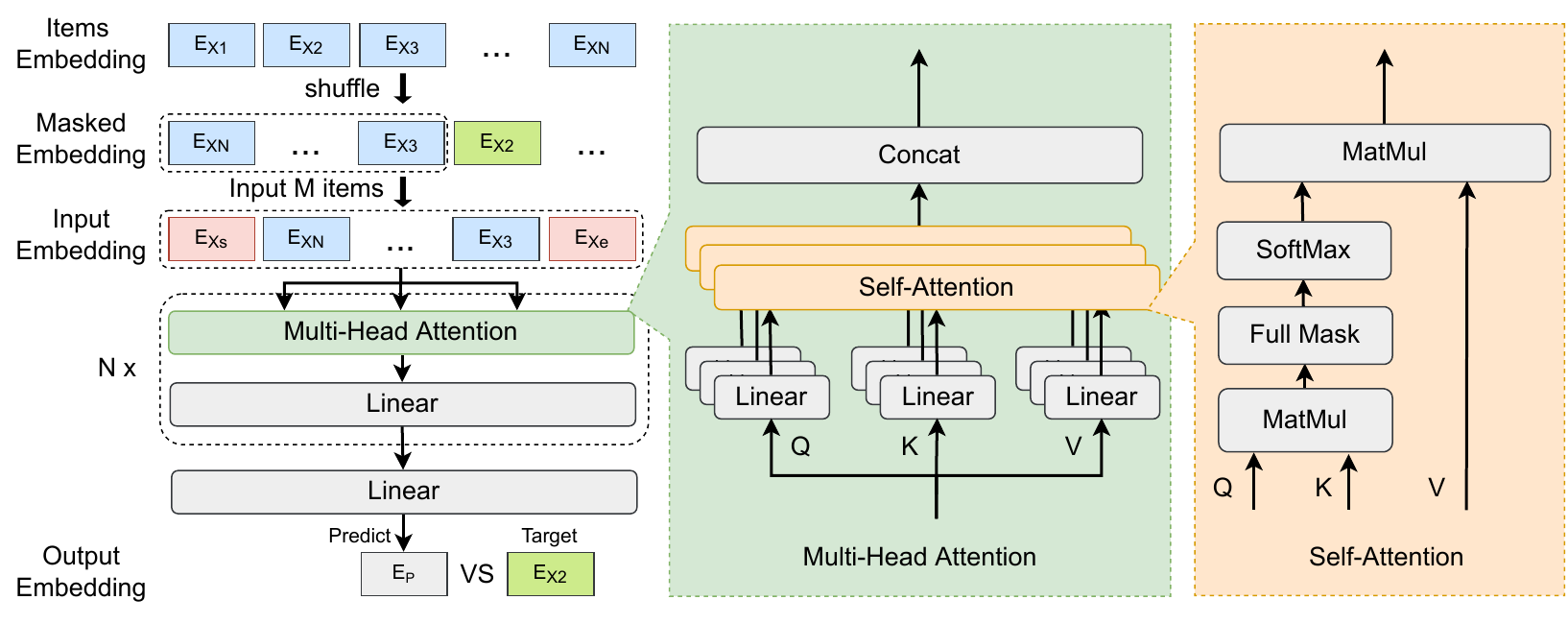}
\caption{\textbf{FBT: Flexible Bidirectional Transformer.} We randomly sample $M\in[0,N]$ items from the total item number $N$ of items (in a scene) as input set, and the $(M+1)_{th}$ item not in the input set as output target. We put the scene embedding at the beginning of input set, and take the scene embedding as the start token $E_{X_s}$. We set a zero vector as the end token $E_{X_e}$. The output will be the end token when the input set contains all the items in the scene. 
}
\label{fig:fbt}
\vspace{-3mm}
\end{figure*}

\subsection{More Details of Our Approach} % we accidentally ref both FTB and SFID to section 3
\subsubsection{Visual Similarity Learning}

%~\cite{jun2019combination}
For visual similarity learning, we apply a standard ResNet50 followed by a combination of average pooling and generalized mean pooling global descriptors. We average the two pooling global descriptors as the final 2048-dimensional image embedding. This model is trained with Normalized Softmax Loss and Soft-margin Triplet Loss for retrieval task.

\paragraph{Normalized Softmax Loss~\cite{zhai2018classification}} The core idea %proposed in this work 
is to use the standard cross-entropy loss 
%employed for training classification models, 
for metric learning. 
The main insight is that there is no need to do exhaustive all-pair comparisons
%between individual training samples, 
as would be ideally done with contrastive and triplet losses.
%if it were feasible. 
Instead, the comparison can be done only to “class proxies” which can be thought of as class representatives or class centroids.

\paragraph{Soft-margin Triplet Loss~\cite{hermans2017defense}}
%Hinge function in Triplet Loss to avoid correcting “already correct” triplets. 
In Triplet loss, the hinge function can help avoid over correcting.
But for similarity learning, it can be beneficial to over correct.
%to pull together samples from the same style as much as possible. 
In this loss, the hinge function is replaced with a softplus function $ln(1 + exp(\cdot))$.
%For this purpose, it is possible to replace the hinge function by a smooth approximation using the softplus function: $ln(1 + exp(\cdot))$, for which numerically stable implementations are commonly available. 
The softplus function has similar behavior to the hinge, but it decays exponentially.
%instead of having a hard cut-off, hence refer to it as the soft-margin formulation. For this work, we employ the soft-margin version which avoids an extra hyperparameter for the hard margin. 

\subsubsection{Complementarity Learning with Flexible Bidirectional Transformer}

As shown in Figure~\ref{fig:fbt}, we randomly mask out $[0,M], M \leq N$ (where N is the total number of items in the scene) items in the sequence as input. Then the model predicts the next $(M+1)th$ item in the sequence. 
We put the scene embedding at the beginning of input set as the start token. 
We set a zero vector as the end token. 
The output will be the end token when the input set contains all the products in the scene.

\subsubsection{SFID}
%For most existing methods, researchers use FITB as the main evaluation metric for CIR task. 
%FITB is a retrieval-based metric, which works fine for single item complementary retrieval. But when it comes to multi-item complementary retrieval, it's difficult for FITB to consider the overall optimality within the entire group, while maintaining the style compatibility. 
%We propose a visual-based metric that can measures the stylistic compatibility of overall item set. 
For better visual sets compatibility evaluation, we propose the SFID metric.
The goal is to propose a metric that correlates the compatibility of a generated set with the ground-truth design set of objects. How to compose a set of object images to form an image is a critical part of the SFID design.
%where $\mu$ and $\sigma$ refer to the mean and covariance of the provided data, $tr$ refers to the trace linear algebra operation.
%\subsubsection{SFID: Style-FID}

We identify three major factors impacting the effectiveness of the score when composing the set image, i.e. background, position, and size of the objects. 
In the ideal scenario, the score computed from the composed images of the generated sets and the ground truth sets should be as low as possible. To compute the metric, we compute images from the generated and groundtruth object sets. And for all the experiments, we set the maximum number of items to be included in one composed image to 5 with the observation of number of items exist in most of the scene images.

%We conduct a set of experiments based on these three major factors.
From the experiments, we find that the background images have larger impact than the compatibility between two sets of items (shown first set in Table~\ref{tab:sfid}).
And varying the size produces negative effect on the score (shown in the second set in Table~\ref{tab:sfid}). 
But, according to the third set in Table~\ref{tab:sfid}, location has small impact on the score compared to other two factors.
Thus, we conclude to use white background, fix the size and randomly position the items when composing the set image when computing the SFID score.
Our perceptual study~\ref{sec:user_study} shows strong positive correlation ($\text{p-value}=0.7$) between SFID and human perception.

After studying the major factors, we conclude fixing the size and randomly place the item images on a white background (shown in Figure~\ref{fig:sfid}a) is better than other setting  with bbox (from the scene image) (shown in Figure~\ref{fig:sfid}b).

\begin{table}
%\vspace{-2mm}
\centering
\resizebox{1.0\columnwidth}{!}{
\begin{tabular}{c c c}
\toprule
Factor & Pair & SFID $\downarrow$    \\
\midrule 
\multirow{3}{*}{Background} & bbox GT VS (bbox GT+scene image) & 98.1  \\
 & bbox Neg VS (bbox GT+scene image) & 107.5  \\
 & bbox GT VS bbox Neg & $75.2$   \\
\midrule 
\multirow{3}{*}{Size} & bbox GT VS non-bbox GT & 104.9  \\
 & bbox GT VS bbox Neg & $75.2$   \\
 & non-bbox GT VS non-bbox Neg & $49.0$ \\
\midrule 
Position & non-bbox GT VS random non-bbox GT & 35.1  \\
\bottomrule 
\end{tabular}
}
\vspace{1mm}
\caption{{\bf Composing Method Comparison.} Here bbox means using the bbox from the original scene image to place in the composed image and non-bbox means using a fixed size. GT is the ground truth set and Neg is the randomly chosen set. 
%The goal in this experiment is to identify a way to compose the set image so that the score correlates with the difference in terms of compatibility. 
With the results, we find using white background and a fixed size for all the items and random placement in the composed set of images produces the lowest score.
}
\label{tab:sfid}
\end{table}

\subsection{More Qualitative Evaluation Results}
%\TODO{How many sets for each scenario? right now the number is 6.}
%To generate the results, 
In this qualitative evaluation, we give our ICAR model: 1) the sampled scene images from the STL-home and STL-fashion dataset test-split; 2) a set of item categories; 3) a pool of cross-domain item images.
%Then we specify the item categories. 
We test our model on different input setup: different sets of item categories and seen or unseen cross-domain item pool. 

\begin{figure*}[!h]
\centering
\includegraphics[width=1.0\textwidth]{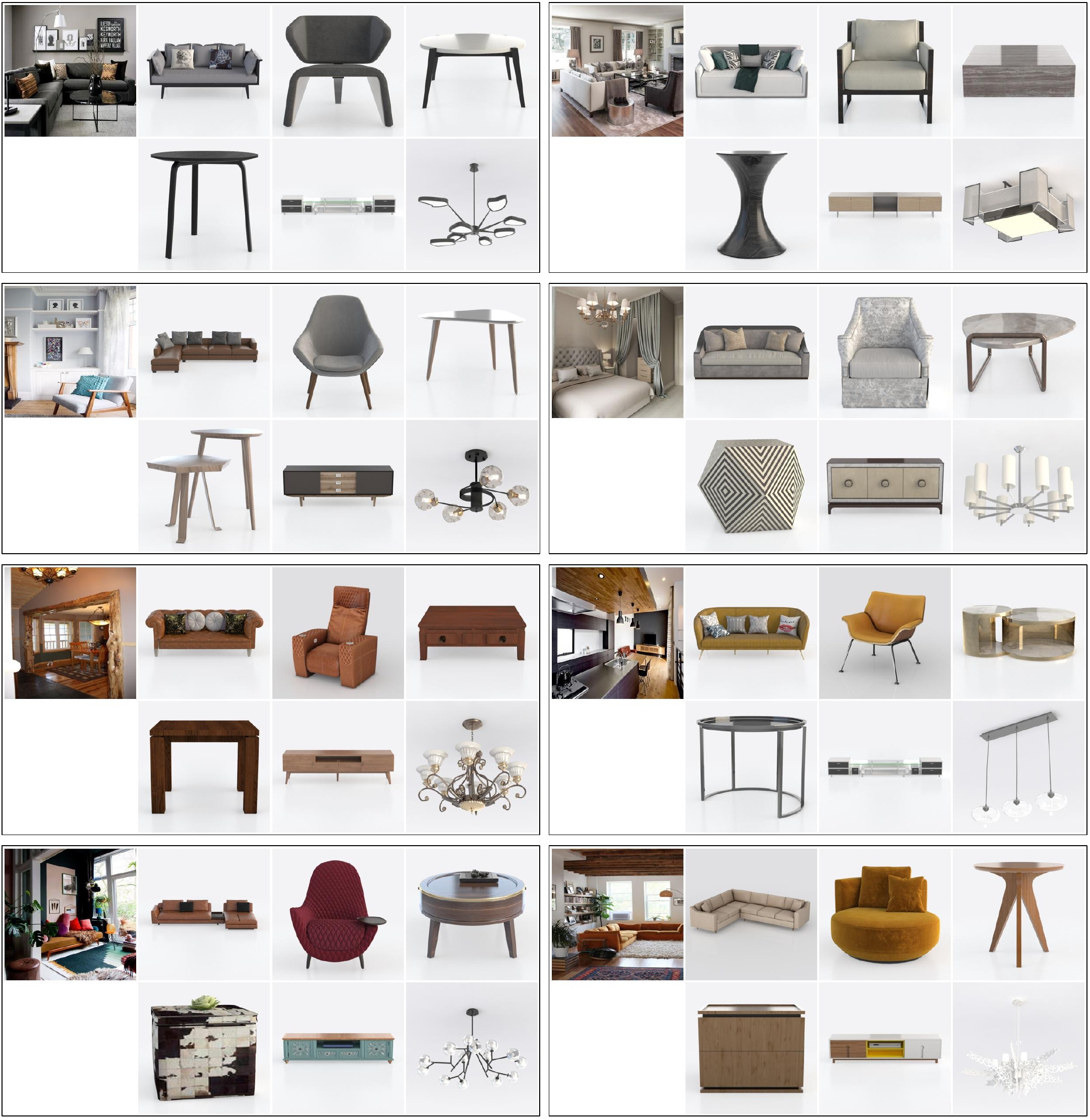}
\caption{\textbf{More Visualization for living room.} In this example, given an inspirational home scene image (sampled from STL-home as shown in column 1, 5) with a pool of products (from 3D-FRONT). Our model auto-regressively retrieves a set of stylistically
compatible items (column 2, 3, 4, 6,7,8 in the picture).
}
\label{fig:living}
\end{figure*}

\begin{figure*}[!h]
\centering
\includegraphics[width=1.0\textwidth]{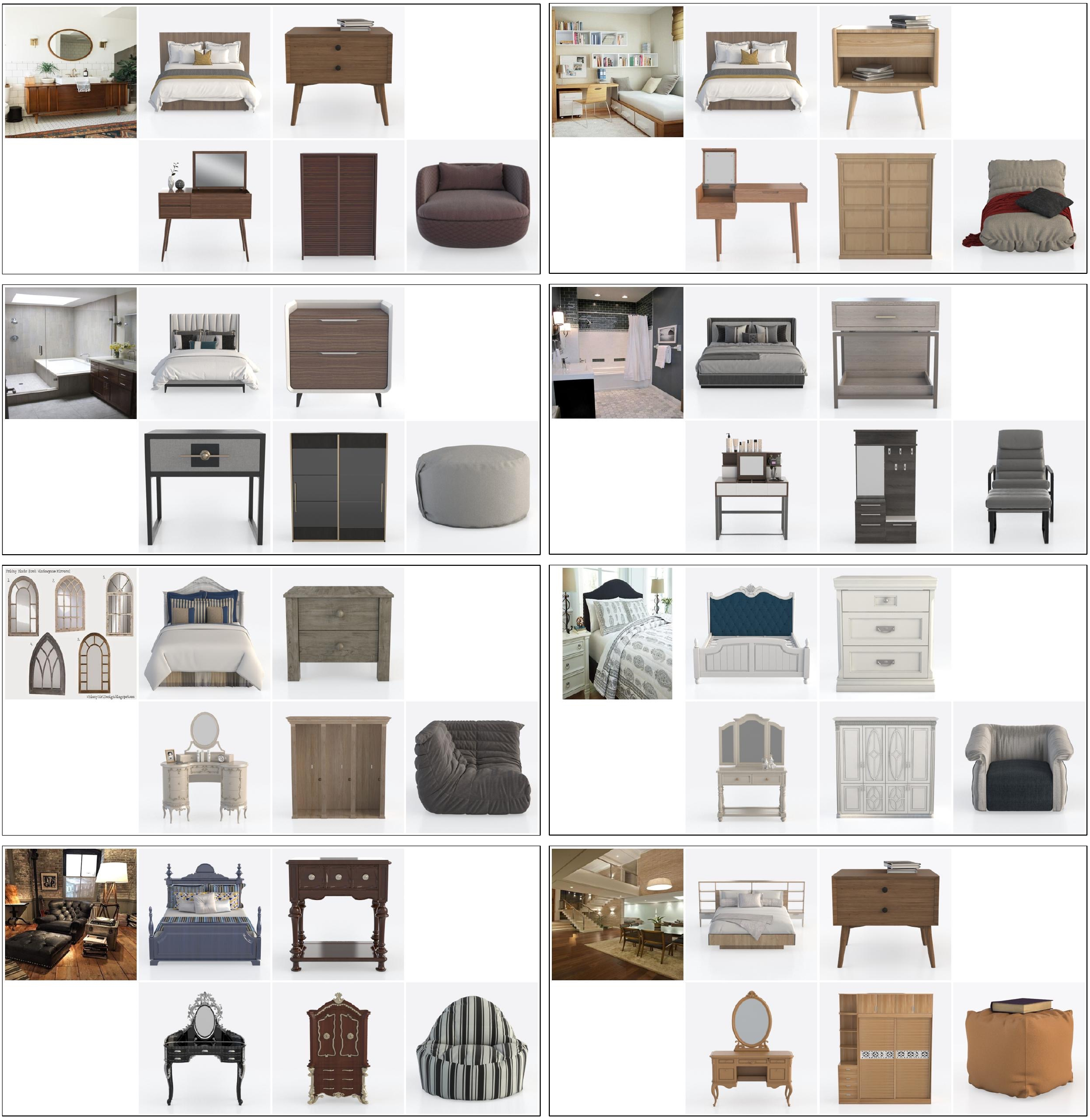}
\caption{\textbf{More Visualization for bedroom.} In this example, given an inspirational home scene image (sampled from STL-home as shown in column 1, 5) with a pool of products (from 3D-FRONT). Our model auto-regressively retrieves a set of stylistically
compatible items (column 2, 3, 4, 6,7,8 in the picture).
}
\label{fig:bedroom}
\end{figure*}

\begin{figure*}[th]
\centering
\includegraphics[width=1.0\textwidth]{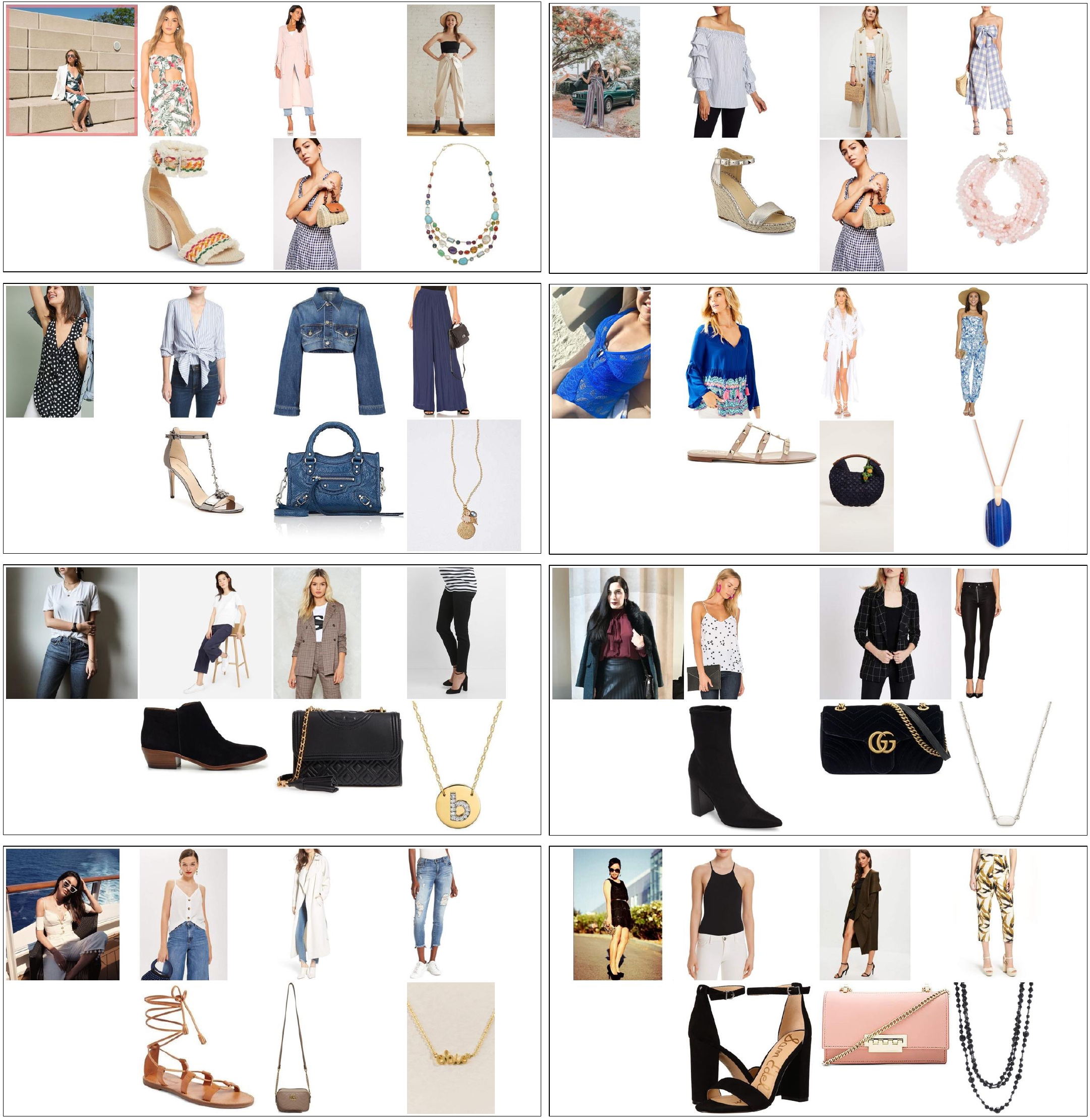}
\caption{\textbf{More Visualization for Fashion.} In this example, given an inspirational fashion scene image (sampled from STL-fashion as shown in column 1, 5) with a pool of products (from STL-fashion). Our model auto-regressively retrieves a set of stylistically
compatible items (column 2, 3, 4, 6, 7, 8 in the picture).
}
\label{fig:fashion}
\end{figure*}

For home recommendation, we test living room and bedroom scene generation.
For living room scene (shown in Figure~\ref{fig:living}), 
%as shown in Figure~\ref{fig:living}, 
we set sofa, chair, coffee table, end table, tv-stand, and ceiling light as given categories. 
For bedroom scene (shown in Figure~\ref{fig:bedroom}), 
%as shown in Figure~\ref{fig:bedroom}, 
we use bed, nightstand, vanity, dresser, and chair as given categories. 
For fashion outfit generation (shown in Figure~\ref{fig:fashion}), 
%as shown in Figure~\ref{fig:fashion}, 
we use top, coat, pant, shoes, handbag, and necklace as given categories. 
Both living room and bedroom scene generation, we test our model's ability in generalizing to unseen domain (3D-FRONT~\cite{fu20213d} synthetically rendered).

\begin{figure*}[!h]
\centering
\includegraphics[width=1.0\textwidth]{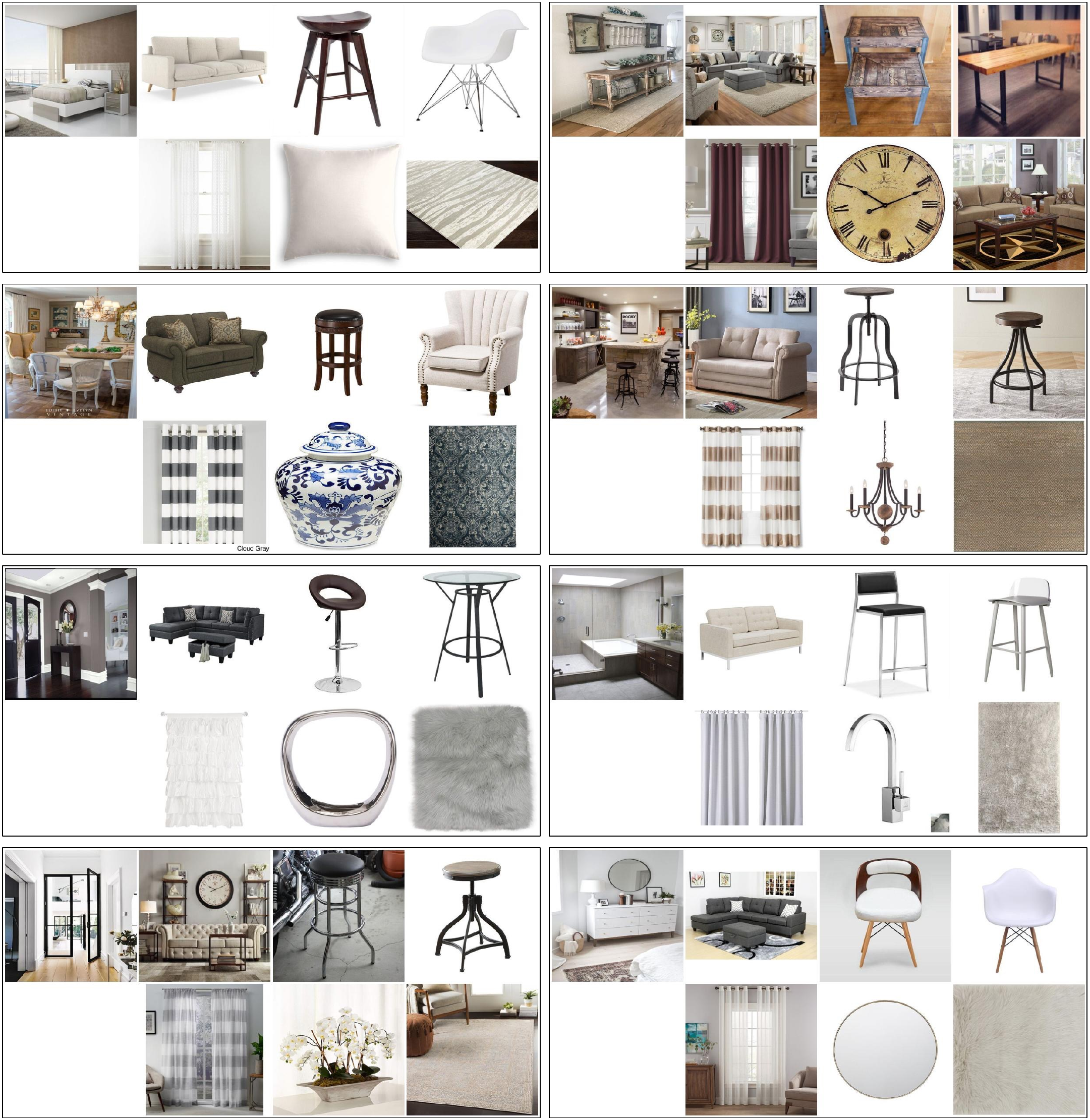}
\caption{\textbf{More Visualization for STL-home.} In this example, given an inspirational home scene image (sampled from STL-home as shown in column 1, 5) with a pool of products (from STL-home). Our model auto-regressively retrieves a set of stylistically compatible items (column 2, 3, 4, 6,7,8 in the picture).
}
\label{fig:stl_home}
\end{figure*}

As shown in Figure~\ref{fig:living} and Figure~\ref{fig:bedroom}, our model can generate stylistically compatible object set from unseen domain.
%Living room and bedroom scene examples are given an inspirational scene image from STL and recommendation products pool from 3D-FRONT. 
For fashion examples, we test our model on the STL white-background object image pool. 
For home recommendation (shown in Figure~\ref{fig:stl_home}), 
we also show results of generating from the STL white-background object image pool. As shown in Figure~\ref{fig:fashion} and Figure~\ref{fig:stl_home}, the recommended sets in both scenarios are style compatible.
%recommendation products are from STL. 

In summary, 
%no matter scene images and product images come from the same domain or not, 
our model can generate high quality style compatible items given different set of categories and can generalize to unseen domains.
\begin{figure}[!t]
\centering
\includegraphics[width=1.0\columnwidth]{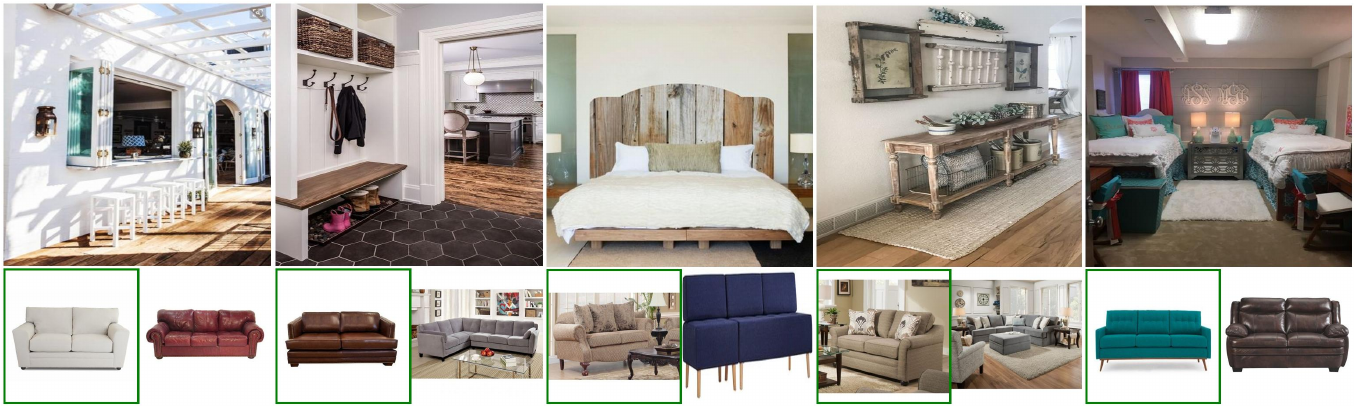}
\caption{\textbf{Binary FITB Results.} We show the FITB results on the STL-home datasets when there are two candidates (second row). Our model chooses the item in the green box. 
}
\label{fig:fitb}
\vspace{-3mm}
\end{figure}
In addition, we conduct the binary FITB (1 negative candidate) experiments~\cite{kang2019complete}, as shown in Figure~\ref{fig:fitb}, our model chooses the more stylistically compatible item in the green box.

%We have two modes to compose the item images: bbox--paste the item image based on the bounding box information; non-bbox--scale the item image into the same size and paste the item image in a grid manner. 
% \subsection{Category Supervision}
% As shown in Table~\ref{tab:category}, we conducted ablation study about category embedding on DeepRooms, our category embedding improves the recommendation performance in both setting-- given the category or predicting the category.


\end{document}